\documentclass[10pt,journal,compsoc]{IEEEtran}
\ifCLASSOPTIONcompsoc
  \usepackage[nocompress]{cite}
\else
  \usepackage{cite}
\fi
\usepackage{graphicx}
\usepackage{amsmath}
\usepackage{amssymb}
\usepackage{booktabs}

\usepackage{capt-of}
\usepackage{times}
\usepackage{epsfig}
\usepackage{graphicx}
\usepackage{amsmath}
\usepackage{amssymb}
\usepackage{xcolor}
\usepackage{blindtext}
\usepackage{overpic}
\usepackage{transparent}
\usepackage{booktabs}
\usepackage{multirow}
\usepackage[super]{nth}
\usepackage[format=plain,labelformat=simple,labelsep=period,font=small,skip=4pt,compatibility=false]{caption}
\usepackage[font=footnotesize,skip=2pt]{subcaption}
\usepackage{makecell}

\makeatletter
\newcommand\notsotiny{\@setfontsize\notsotiny\@vipt\@viipt}

\makeatother

\usepackage{etoolbox}
\usepackage[binary-units]{siunitx}
\robustify\bfseries
\DeclareSIUnit{\inch}{inch}

\newcolumntype{C}[1]{>{\centering\arraybackslash}p{#1}}

\usepackage{xspace}
\makeatletter
\DeclareRobustCommand\onedot{\futurelet\@let@token\@onedot}
\def\@onedot{\ifx\@let@token.\else.\null\fi\xspace}
\def\eg{\emph{e.g}\onedot} 
\def\ie{\emph{i.e}\onedot} 
 
 \def\vs{\emph{vs}\onedot}
 
\def\etal{\emph{et al}\onedot}
\makeatother

\usepackage{pifont}
\definecolor{lightgray}{rgb}{0.9, 0.9, 0.9}
\definecolor{lgray}{rgb}{0.66, 0.66, 0.66}
\ifCLASSINFOpdf
\else
\fi
\begin{document}
%
\title{Multimodal-driven Talking Face Generation via a Unified Diffusion-based Generator}
%
%
%
%

\author{Chao~Xu,
        Shaoting~Zhu,
        Junwei~Zhu,
        Tianxin~Huang,
        Jiangning~Zhang,
        Ying~Tai,
        Yong~Liu
\IEEEcompsocitemizethanks{\IEEEcompsocthanksitem C. Xu, S. Zhu, T. Huang and Y. Liu are with APRIL Lab, Zhejiang University, Hangzhou, China (e-mail: 21832066@zju.edu.cn; zhust@zju.edu.cn; 21725129@zju.edu.cn; yongliu@iipc.zju.edu.cn).
\IEEEcompsocthanksitem J. Zhu, J. Zhang, and Y. Tai are with YouTu Lab, Tencent, Shanghai, China (e-mail: junweizhu@tencent.com; vtzhang@tencent.com; yingtai@tencent.com).
\IEEEcompsocthanksitem Y. Liu is the corresponding author.
}
\thanks{Manuscript received April 19, 2005; revised August 26, 2015.}}

%
%

\markboth{Journal of \LaTeX\ Class Files,~Vol.~14, No.~8, August~2015}%
{Shell \MakeLowercase{\textit{et al.}}: Bare Demo of IEEEtran.cls for Computer Society Journals}
%



\IEEEtitleabstractindextext{%
\begin{abstract}

Multimodal-driven talking face generation refers to animating a portrait with the given pose, expression, and gaze transferred from the driving image and video, or estimated from the text and audio. However, existing methods ignore the potential of text modal, and their generators mainly follow the source-oriented feature rearrange paradigm coupled with unstable GAN frameworks. In this work, we first represent the emotion in the text prompt, which could inherit rich semantics from the CLIP, allowing flexible and generalized emotion control. We further reorganize these tasks as the target-oriented texture transfer and adopt the Diffusion Models. More specifically, given a textured face as the source and the rendered face projected from the desired 3DMM coefficients as the target, our proposed Texture-Geometry-aware Diffusion Model decomposes the complex transfer problem into multi-conditional denoising process, where a Texture Attention-based module accurately models the correspondences between appearance and geometry cues contained in source and target conditions, and incorporate extra implicit information for high-fidelity talking face generation. Additionally, TGDM can be gracefully tailored for face swapping. We derive a novel paradigm free of unstable seesaw-style optimization, resulting in simple, stable, and effective training and inference schemes. Extensive experiments demonstrate the superiority of our method.
    
\end{abstract}

}

\maketitle

\IEEEdisplaynontitleabstractindextext

%
\IEEEpeerreviewmaketitle

\IEEEraisesectionheading{
\section{Introduction}\label{sec:introduction}}

Talking face generation aims to synthesize talking video from the source face according to the given emotion, mouth movement, and head rotation, which is relevant to several applications, including video production and virtual avatars. Multimodal information could guide the animation in the real scenario, such as text, audio, image, and video. 

\begin{figure}[t!]
	\centering
	\includegraphics[width=0.48\textwidth]{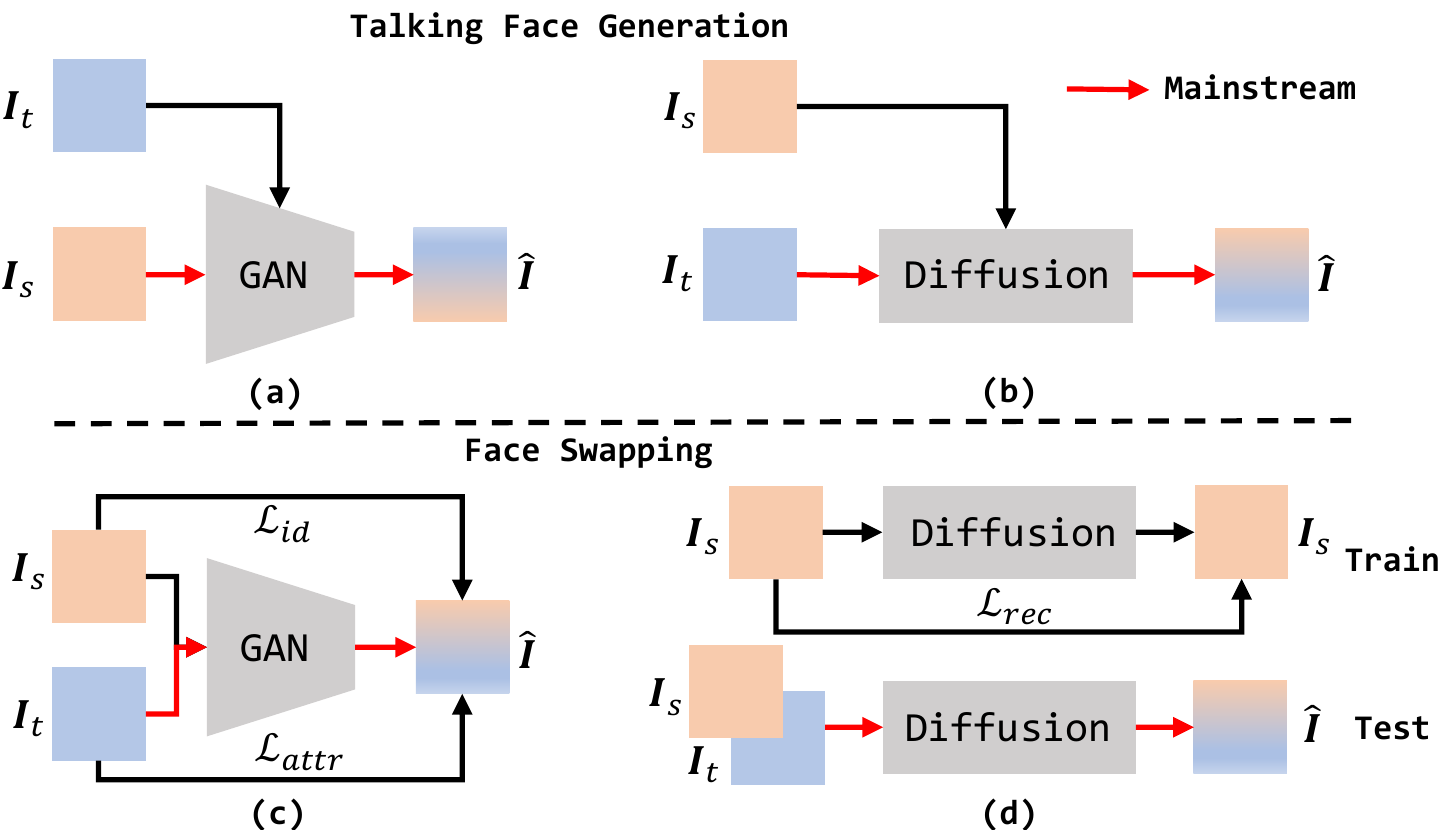}
	\caption{The top part shows the proposed generator in talking face generation. Recent mainstream methods (a) are source-oriented rearranging operation, while our proposed pipeline (b) serve as the target-oriented texture transfer task. The bottom part shows the extended application of the generator in face swapping. Current methods (c) suffer from unstable training introduced by GANs and seesaw-style losses optimized difficultly. By contrast, we propose a new paradigm (d) with simple reconstruction loss and stable training while still maintaining comparable results.
	}
	\label{fig:teaser}
\end{figure}

Recently, many attempts~\cite{zhou2021pose, liang2022expressive, siarohin2019first} have achieved significant progress in these tasks, most of them share the same paradigm, \ie, extracting the intermediate structural representation from given conditions first and then manipulating the source face to the desired expression and pose, which mainly follows the source-oriented pipeline, as shown in Fig.~\ref{fig:teaser} (a). Specifically, among these approaches, some image- and video-driven~\cite{kim2021exploiting, zeng2020realistic, burkov2020neural} methods employ AdaIN-based~\cite{huang2017arbitrary} generators that take vectors as input, which inevitably lead to the information loss and fail to preserve the source identity and background. Others~\cite{ren2021pirenderer, siarohin2019first, zhao2022thin, tao2022structure, ji2022eamm} warp the source feature to the target by the explicit motion flows for better visual results, but they appear warping artifacts when the source and driving conditions encompass significant appearance variation. Consequently, the above generators tend to suffer from image degradation when rearranging the source features. Recent audio-driven tasks require more authentic results, subsequent works~\cite{zhou2020makelttalk, zhang2021facial, guo2021ad} adopt identity-specific training but cannot generalize across different persons. Besides, existing pipelines generally adopt GANs~\cite{gan}, and its unstable adversarial min-max objective training process further exacerbates unrealistic textures. Due to these constraints of the generator, each task needs a specific design and is unfriendly for practical applications. Thus, one challenge arises, how to accomplish a robust and stable generator for all driving modals to achieve high-fidelity talking face generation. In addition, existing work~\cite{li2021write} simply reflects the text on the mouth movements, ignoring the potential of text when under the large-scale pre-trained models. Thus another challenge arises, how to sufficiently use the text modal in this task.

To address the above challenges, we first represent the emotion style in the text prompt inspired by the zero-shot CLIP-guided image manipulation, which could inherit rich semantic knowledge and allow flexible emotion control, \ie, unseen emotions could be specified using the text description and precisely reflected on the synthesized faces. Furthermore, to unify the multimodal-driven tasks into the same generator, we frame the talking face generation as a target-oriented texture transfer, instead of the source-oriented feature rearrange, and adopt a multi-conditional diffusion model to avoid unstable training of GANs, termed Texture-Geometry-aware Diffusion Model (TGDM), as shown in Fig.~\ref{fig:teaser} (b). In particular, benefiting from the explainable and disentangled parameter space of 3DMMs~\cite{deng2019accurate}, we combine the texture-related coefficients from the source face with the geometry-related ones from the driving conditions to construct 3D descriptors, which are projected to the image domain and serve as the target pivot. To further supplement source texture to rendered face, we employ cross attention that accurately models the correspondences between source and target appearance. To this end, TGDM is dedicated to transferring the source texture to the target rendered face, which preserves explicit structural information but avoids complex texture deformations. In contrast to recent diffusion-based methods~\cite{bigioi2023speech, shen2023difftalk} that only handle mouth area generation, our approach can generate realistic faces with various expressions and poses.

Considering the characteristics of the TGDM to model complex texture and semantic transfer, we further connect TGDM with another popular task, face swapping, which aims to transfer the source identity to the target face while preserving the 
target attributes. Recent developments are stuck due to unstable GAN-based training schemes and seesaw-style optimization goals. DiffFace~\cite{diffface} first avoids GANs but is still sensitive to identity-related and identity-unrelated hyperparameter settings when sampling. Borrowing the idea from the aforementioned driving framework, we derive a novel paradigm for face swapping built upon the TGDM, which inherits the merits of the diffusion model and requires only reconstruction loss during training, with no extra tricks for sampling either, as shown in Fig.~\ref{fig:teaser} (d).

In summary, we make the following four contributions:
\begin{itemize}

\item We adopt the text modal as the talking face emotion representation, inheriting rich semantics from large-scale pre-trained models, which allows flexible emotion control and unseen emotion generalization.

\item We propose a novel TGDM pipeline based on the multi-conditional diffusion model to afford complex texture and identity transfer, generating high-quality talking face generation for all driven modals.

\item We transfer the TGDM to face swapping task and propose a novel training and inference paradigm that is simple, stable, and effective.

\item Abundant experiments are conducted to demonstrate the superiority of TGDM for several face manipulation tasks over SOTA methods.

\end{itemize}

\section{Related Works}
\subsection{Talking Face Generation}

Face reenactment involves taking the source face and replicating its pose and expression as the target. This can be achieved through two main techniques: instruction-based methods animate the source face instructed by the target structure. Various works~\cite{wu2018reenactgan, huang2020learning, zhang2020freenet, ha2020marionette, chen2020puppeteergan} adopt landmarks and segmentation maps to indicate the facial attribute. Recently, with the success of AdaIN~\cite{huang2017arbitrary}, subsequent works~\cite{zeng2020realistic, zakharov2019few} encode the target attributes in the vectorized information and then inject them into the source face. However, the above methods fail to explicitly indicate the movements between the source and target faces. Subsequently, warping-based methods learn to warp and synthesize the target faces based on the estimated motion fields. These methods~\cite{wiles2018x2face, siarohin2019monkeynet} usually separate motion estimation and warped source face refinement into two stages. The most representative work is FOMM~\cite{siarohin2019first}, which uses relative key-point locations to predict flow fields for source appearance driving. Other follow-up works~\cite{tao2022structure, zhao2022thin} focus on improving the motion flows and warping operation accuracy. Some works~\cite{ren2021pirenderer, zhang2021flow, doukas2021headgan, hong2022depth} introduce 3D information as structure guidance for flow field generation. However, they still suffer from identity degradation under some extreme conditions. Recent UniFace~\cite{xu2022uniface} proposes a unified framework to boost the model's robustness with the help of face swapping.  

Audio-driven talking head synthesis, a special form of face reenactment, aims to create talking videos with lip movements corresponding to the driving audio~\cite{garrido2015vdub, suwajanakorn2017synthesizing, prajwal2020lip}. The traditional approaches could be roughly divided into 2D-based and 3D-based ones. 2D-based methods~\cite{aneja20192d1, biswas20212d2, eskimez20182d3,ji20212d4} generate a series of 2D points on the face based on audio inputs, while 3D-based methods~\cite{chen20203d1, cudeiro20193d2, karras20173d3, richard20213d4, guo2021ad, shen2022learning} use audio to predict expression parameters or facial radiance fields. Some works~\cite{hong2022depth, min2022styletalker, zhou2020makelttalk} further improve geometry learning and take talking style into account. After achieving the intermediate structure, PC-AVS~\cite{zhou2021pose} injects the pose and lip information into the generator by implicit modulation. StyleHeat~\cite{yin2022styleheat} generates high-resolution driven faces with the help of StyleGAN. Besides, emotion is a factor that plays a critical role in realistic animation. MEAD~\cite{wang2020mead} releases a high-quality talking head video dataset with annotations of emotion category and intensity. Subsequent works~\cite{liang2022expressive, ji2022eamm} follow the framework of PC-AVS or FOMM and take emotion as another condition. However, due to the limitations of the generator and unstable GAN-based training, the above methods need to design the specific intermediate representation and generator for each driving modal, thus making it impossible to share the same structure.
Nowadays, some attempts based on diffusion models~\cite{diffusion1,diffusion2} have been made. Stypułkowski \etal~\cite{stypulkowski2023diffused} leverage a pre-trained audio encoder to add audio embeddings during the denoising process. DiffTalk~\cite{shen2023difftalk} and Bigioi \etal~\cite{bigioi2023speech} present crafted conditional diffusion models for generalized talking head synthesis. However, they fail to model the distinct expression and pose variations.

\subsection{Face Swapping}

Face swapping aims to change the target identity according to the given source but keep other facial attributes constant. Early face swap works~\cite{blanz2004exchanging,bitouk2008face,cheng20093d,lin2012face} mainly focus on 3D-based methods but suffer from poor visual quality. Recently, GAN-based~\cite{gan} methods~\cite{deepfake, RSGAN, IPGAN} have made significant progress. Specifically, Faceshifter~\cite{faceshifter} integrates identity and attribute embeddings adaptively from a well-designed learning model. SimSwap~\cite{chen2020simswap} introduces a feature matching loss hoping to preserve more attribute embeddings at the cost of sacrificing identity similarity. Hififace~\cite{wang2021hififace} and FaceInpainter~\cite{faceinpainter} take 3D face descriptor into consideration for better geometry structure of swapped results. 
With the success of StyleGAN~\cite{stylegan, stylegan2}, many works have emerged as a solution for high-resolution face swap. MegaFS~\cite{megafs} first exploits StyleGAN2 as the decoder. The follow-up works~\cite{xu2022high,xu2022region} also adopt the pSp~\cite{richardson2021encoding} framework and design the fusion strategy for better attribute preservation. However, they lack flexibility in application due to the fixed StyleGAN generator. Consequently, some attempts have been made to solve this problem. StyleFace~\cite{styleface} redesigns the StyleGAN2 module and opens parameters for training. StyleSwap~\cite{styleswap} introduces a mask branch and an ID inversion strategy to empower high-fidelity and robust face swapping.
As the diffusion model shows excellent performance in many fields, DiffFace~\cite{diffface} makes the first attempt to apply the diffusion model to the face swapping task. Despite the impressive progress achieved by the above methods, it is still a struggle to fully transfer the face identity from the source face while preserving identity-unrelated attributes of the target images due to seesaw-style training losses. One solution~\cite{gao2021information} is to fully disentangle identity-related and identity-unrelated information, but it is almost impossible in the current implementation scheme. In this paper, we propose a new training paradigm only guided by the reconstruction loss to solve this challenge.

\subsection{Diffusion Model and Multimodal Generation}

 Diffusion models~\cite{diffusion1, diffusion2} are recently proposed generative models that can synthesize high-quality images. They are a type of generative probabilistic model that consists of two steps. Firstly, data is destroyed by successively adding small amounts of Gaussian noise to it over a series of time steps. Secondly, a learning algorithm is trained to recover the data by gradually removing the noise over a series of time steps.  Diffusion models are trained without discriminators, so they are more reliable and robust during training compared to GANs. Additionally, they do not suffer from common issues such as mode collapse or vanishing gradients, which are inevitable in the training process of GANs. After achieving great success in the unconditional generation, diffusion models are adapted to enable conditional generation. Dhariwal \etal~\cite{dhariwal2021adm} introduce classifier-guided diffusion, which forces the produced noise to approach the desired condition. Ho \etal further~\cite{ho2022classifier} develop a Classifier-Free Guidance approach that allows conditional editing without having to pretrain classifiers. Despite these advantages, diffusion models are hindered by their slow sampling speed due to the thousands of times on one sample for complete pixel space-based denoising. To address this issue, Song \etal~\cite{song2020denoising} propose DDIM reduce sample time, and Rombach \etal~\cite{rombach2022high} propose the Latent Diffusion Models (LDMs), which transfer the training and inference processes to a compressed lower-dimension latent space for more efficient computing.
 
Diffusion models have become increasingly popular in multimodal generation incorporated with CLIP~\cite{radford2021learning} due to their ability to generate data with desirable qualities while covering a wide range of distributions. Application fields of the diffusion model vary from text-based image generation~\cite{avrahami2022text1,fan2022text2,ruiz2022text3,saharia2022text4}, text-based video generation~\cite{ho2022video, ho2022imagen, wu2022tune, molad2023dreamix, zhou2020makelttalk}, text-based audio generation~\cite{huang2023make, liu2023audioldm}, text-based 3D representation generation~\cite{poole2022dreamfusion, xu2022dream3d, li2022diffusion}, and many others. In this paper, we build our framework on the diffusion model and focus on animating the source face by multimodal geometry guidance, \ie, text, audio, image, and video, reflecting on facial expression and pose.

\section{Preliminaries}

\subsection{Denoising Diffusion Probabilistic Models (DDPMs)}
DDPMs follow the idea of latent variable models that consist of a forward diffusion process and a reverse diffusion process. Specifically,
a diffusion process gradually adds noise to the data sampled from the target distribution $\boldsymbol{x}_0 \sim q(\boldsymbol{x}_0)$ as a Markov chain. Each step $q\left(\boldsymbol{x}_t \mid \boldsymbol{x}_{t-1}\right)$ (for $t \in \left\{0, \dots, T\right\}$) is defined as a Gaussian distribution with a fixed or learned variance schedule $\beta_t \in(0,1)$:
\begin{equation}
    q\left(\boldsymbol{x}_t \mid \boldsymbol{x}_{t-1}\right):=\mathcal{N}\left(\sqrt{1-\beta_t} \boldsymbol{x}_{t-1}, \beta_t \mathbf{I}\right).
\end{equation}
By the Bayes' rules and Markov property, the latent variable $\boldsymbol{x}_t$ can be expressed as:
\begin{equation}
    q\left(\boldsymbol{x}_t \mid \boldsymbol{x}_0\right)=\mathcal{N}\left(\boldsymbol{x}_t ; \sqrt{\bar{\alpha}_t} \boldsymbol{x}_0,\left(1-\bar{\alpha}_t\right) \mathbf{I}\right),
    \label{eq:addnoise}
\end{equation}
where $\bar{\alpha}_t=\prod_{s=1}^t \alpha_s$, and $\alpha_t = 1 - \beta_t$. Then, the reverse process $q\left(\boldsymbol{x}_{t-1} \mid \boldsymbol{x}_{t}\right)$ can be parametrized by another Gaussian transition:
\begin{equation}
    p_\theta\left(\boldsymbol{x}_{t-1} \mid \boldsymbol{x}_t\right):=\mathcal{N}\left(\boldsymbol{x}_{t-1} ; \mu_\theta\left(\boldsymbol{x}_t, t\right), \sigma_\theta\left(\boldsymbol{x}_t, t\right)\right),
\end{equation}
where $\mu_\theta\left(\cdot\right)$ and $\sigma_\theta\left(\cdot\right)$ are predicted by the trained deep neural networks $\boldsymbol{\epsilon}_\theta$, which is optimized under the objective $\mathbb{E}_{\boldsymbol{x}, \boldsymbol{\epsilon} \sim \mathcal{N}(0,1), t}\left[\left\|\boldsymbol{\epsilon}-\boldsymbol{\epsilon}_\theta\left(\boldsymbol{x}_t, t\right)\right\|_2^2\right]$. Thus, given $\boldsymbol{x}_t$, $\boldsymbol{x}_{t-1}$ can be sampled by using:
\begin{equation}
    \boldsymbol{x}_{t-1}=\frac{1}{\sqrt{1-\beta_t}}\left(\boldsymbol{x}_t-\frac{\beta_t}{\sqrt{1-\alpha_t}} \boldsymbol{\epsilon}_\theta\left(\boldsymbol{x}_t, t\right)\right)+\sigma_t \boldsymbol{z},
\end{equation}
where $\boldsymbol{z} \in \mathcal{N}\left(0, \mathbf{I}\right)$. Furthermore, according to~\cite{song2020denoising}, $\boldsymbol{x}_0$ can be approximate derived by $x_t$ and $\boldsymbol{\epsilon}_\theta\left(\boldsymbol{x}_t, t\right)$:
\begin{equation}
    \hat{\boldsymbol{x}_0}:=\frac{\boldsymbol{x}_t-\sqrt{1-\alpha_t} \epsilon_\theta\left(\boldsymbol{x}_t, t\right)}{\sqrt{\alpha_t}}.
    \label{eq:noised}
\end{equation}
This facilitates the use of pixel-level and perceptual losses during the training stage in Sec.~\ref{sec:TGDM}.

\subsection{3D Morphable Models (3DMMs)}
Recent methods estimate the 3D face descriptors of 2D images by optimizing a neural network to extract 3D parameters from a face image. Thus we follow the previous work D3DFR~\cite{deng2019accurate} that adopts ResNet50 as the backbone to predict 3DMM coefficients, which consists of identity $\boldsymbol{\alpha} \in \mathbb{R}^{80}$, expression $\boldsymbol{\beta} \in \mathbb{R}^{64}$, texture $\boldsymbol{\delta} \in \mathbb{R}^{80}$, illumination $\boldsymbol{\gamma} \in \mathbb{R}^{27}$, and pose $\boldsymbol{p} \in \mathbb{R}^6$. Note that the original 3DMM fails to control gaze direction, we explicitly model the gaze like~\cite{park2018learning}, providing the normalized direction vector from the center of the eye to the pupil in four dimensions $\boldsymbol{\omega} \in \mathbb{R}^{4}$. Therefore, given an input face $\boldsymbol{I}$, the output coefficients $\boldsymbol{\rho} \in \mathbb{R}^{261}$:
\begin{equation}
  \begin{aligned}
    \boldsymbol{\rho} = \boldsymbol{\mathcal{D}}(\boldsymbol{I}) = \left\{\boldsymbol{\alpha}, \boldsymbol{\beta}, \boldsymbol{\delta}, \boldsymbol{\gamma}, \boldsymbol{p}, \boldsymbol{\omega}\right\}.
  \end{aligned}
\end{equation}
With 3DMM, the 3D shape $\mathbf{S}$ and albedo texture $\mathbf{T}$ could be parameterized as:
\begin{equation}
  \begin{aligned}
    \mathbf{S} &= \bar{\mathbf{S}} + \mathbf{B}_{id} \boldsymbol{\alpha} + \mathbf{B}_{exp} \boldsymbol{\beta}, \\
    \mathbf{T} &= \bar{\mathbf{T}} + \mathbf{B}_{t} \boldsymbol{\delta},
  \end{aligned}
\end{equation}
where $\bar{\mathbf{S}}$ and $\bar{\mathbf{T}}$ denote the mean face shape and albedo texture. $\mathbf{B}_{id}$, $\mathbf{B}_{exp}$, and $\mathbf{B}_{t}$ are the bases of identity, expression, and texture computed via PCA. We project the reconstructed 3D face onto the 2D image plane with a differentiable renderer $\mathcal{\boldsymbol{R}}$ according to its illumination $\boldsymbol{\gamma}$ and pose $\boldsymbol{p}$:
\begin{equation}
  \begin{aligned}
    \boldsymbol{I}_{3d} = \boldsymbol{\mathcal{R}}(\mathbf{S}, \mathbf{T}, \boldsymbol{\gamma}, \boldsymbol{p}).
  \end{aligned}
\end{equation}
We naturally choose the rendered image $\boldsymbol{I}_{3d}$ as the intermediate geometry condition in Sec.~\ref{sec:MGC} due to its several appealing properties: 1) Compared with other structural representations, \eg, landmarks and segmentation maps, 3DMMs provide an explainable and disentangled parameter space, which enables direct recombine corresponding factors when conducting the specific face manipulation task. Besides, mapping other cues to 3DMMs is much easier since no additional spatial information is required. 2) Rendered face images provide more detailed semantic and explicit geometry than vectorized parameters, thus reducing the training difficulty. We conduct extensive experiments in Sec.~\ref{sec:5.2}.

\begin{figure*}[t!]
	\centering
	\includegraphics[width=0.85\textwidth]{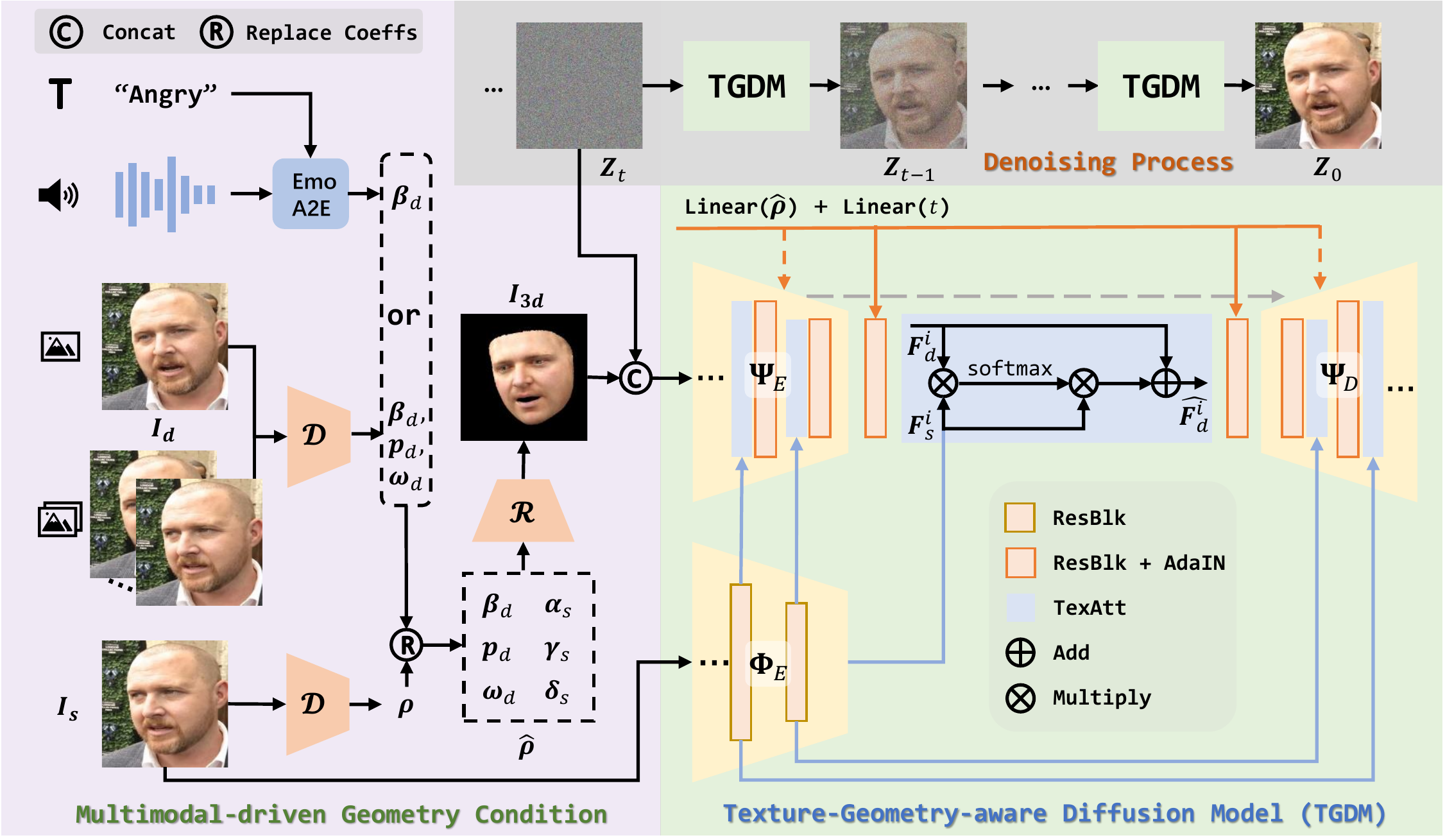}
	\caption{Overview of the proposed method for multimodal-driven talking face generation. Given multimodal conditions, we first represent them in geometry-related 3DMM coefficients and then project them to the rendered face $\boldsymbol{I}_{3d}$, which serves as the intermediate structural representation. Deviating from previous works, we employ text as the emotion signal for flexible and generalized control. To obtain realistic faces, we develop a multi-conditional diffusion model that learns geometry prior from $\boldsymbol{I}_{3d}$ by simply concatenated input, transfers source appearance from $\boldsymbol{I}_s$ by cross attention, and supplements implicit identity and geometry information by AdaIN.}
	\label{fig:pipeline_reenact}
\end{figure*}

\section{Method}
\subsection{Overview}
As shown in Fig.~\ref{fig:pipeline_reenact}, multimodal-driven talking face generation aims to produce realistic videos according to given source identity and multimodal geometry conditions, \ie, text, audio, image, and video. In Multimodal-driven Geometry Condition, we employ rendered faces projected from 3DMMs as the intermediate structural representation. To further exploit the potential of text in this task, we represent the emotion style in text prompts, which could inherit rich semantics from the large-scale pre-trained models for flexible and generalized emotion control (Sec.~\ref{sec:MGC}). To enable multimodal conditions to share the same generator, we propose a powerful paradigm, termed Texture-Geometry-aware Diffusion Model (TGDM), which is based on the multi-conditional diffusion model, allowing complex texture transfer for high-fidelity face generation, and avoids unstable GAN-based training (Sec.~\ref{sec:TGDM}). 
Finally, we extend TGDM to face swapping and derive a new paradigm for stable and effective training and inference (Sec.~\ref{sec:swap}). In the following, we will supply more details.

\subsection{Multimodal-driven Geometry Condition}
\label{sec:MGC}

\noindent\textbf{Image-driven and Video-driven Conditions.} For image driving, we combine the appearance-related 3DMM coefficients (identity, texture, and illumination) from the source image $\boldsymbol{I}_s$ with the motion-related coefficients (expression, pose, and gaze) from the driving image $\boldsymbol{I}_d$ to construct the desired 3D face descriptors $\boldsymbol{\hat{\rho}} = \left\{\boldsymbol{\alpha}_s, \boldsymbol{\beta}_d, \boldsymbol{\delta}_s, \boldsymbol{\gamma}_s, \boldsymbol{p}_d, \boldsymbol{\omega}_d\right\}$, along with its rendered face $\boldsymbol{I}_{3d}$ as the geometry conditions. For video driving $\left\{\boldsymbol{I}_{1,d}, \boldsymbol{I}_{2,d}, \cdots, \boldsymbol{I}_{N,d}\right\}$, we can treat them as isolated $N$ images for processing. However, parameters from a single input frame will cause jitter and instability in the final generated video due to the inevitable prediction errors between consecutive frames. To alleviate this problem, like ~\cite{ren2021pirenderer}, we introduce a windowing strategy for better temporal consistency, \ie, the parameters of the adjacent frames are also used as descriptors of the central frame to smooth the motion trajectory. In practice, the coefficients of a window with continuous frames $\boldsymbol{\hat{\rho}} \equiv \boldsymbol{\hat{\rho}}_{i-k:i+k}$ and the rendered frame of the central frame as the geometry conditions, where $k$ is the radius of the window and set to 1 experimentally.

\noindent\textbf{Audio-driven and Text-driven Conditions.} Audio-driven talking face generation is expected to maintain lip movements synchronized with input speech contents and synthesize natural facial motion simultaneously. Consequently, it raises two challenges, one is precise audio-to-lip mapping, and the other is highly temporal consistent. Unlike previous works that adopt LSTM~\cite{hochreiter1997long, ren2021pirenderer} or GRU~\cite{lu2021live, cho2014learning} to autoregressively deduce expression coefficients, we adopt the non-autoregressive Transformer~\cite{vaswani2017attention} to capture the short- and long-term audio context and provide the sequence-level representations for more accurate and temporal-coherent coefficients regression. Besides, emotion style also plays a crucial role in generating vivid talking face. For emotion representation, the one-hot coding~\cite{wang2020mead} is in a fixed pattern and fails to convey the semantics cues contained in the label, while recent methods~\cite{ji2022eamm, liang2022expressive} extract emotion embedding from given images and audio, lacking generalizing to unseen styles due to the limited semantics. In contrast, we represent the emotion style in the text prompt and borrow help from CLIP to deliver the semantic cues. Thus our method inherits rich semantic knowledge and convenient interaction after various emotion styles are encoded by CLIP.

\begin{figure}[t!]
	\centering
	\includegraphics[width=0.38\textwidth]{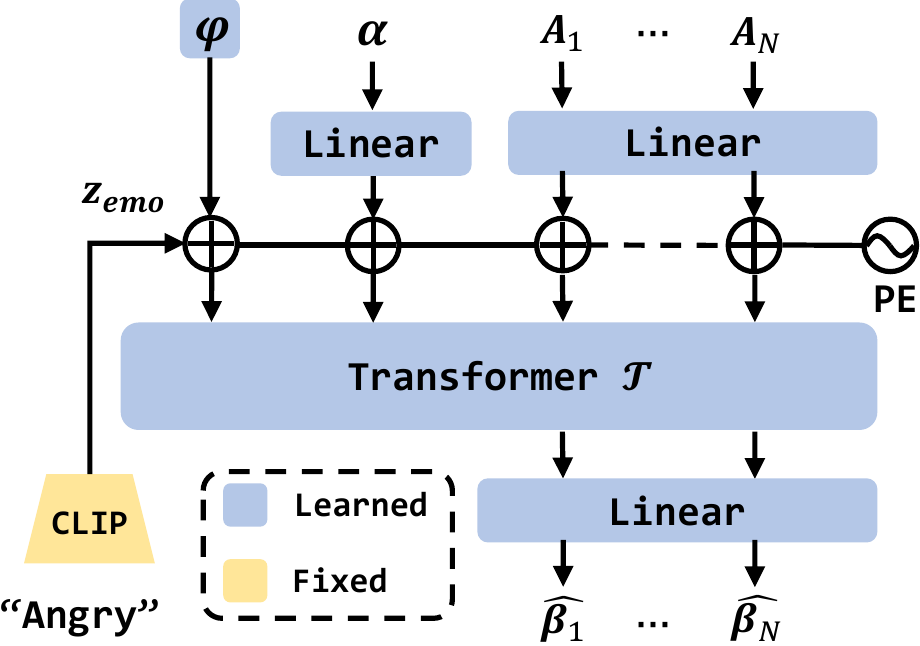}
	\caption{The architecture of EmoA2E module. Removing the emotion embedding $\boldsymbol{z}_{emo}$ and intensity $\boldsymbol{\varphi}$ inputs, this structure is used for emotion-free audio-to-expression learning.
	}
	\label{fig:audio2exp}
\end{figure}

To this end, we propose the Emotional Audio to Expression (EmoA2E) module. Specifically, as shown in Fig.~\ref{fig:audio2exp}, the Mel-frequency Cepstral Coefficients (MFCC) clips $\boldsymbol{A} = \left\{\boldsymbol{A}_1, \dots,\boldsymbol{A}_N\right\}$ provide the cues of lip movement, a non-learnable extended token takes identity information $\boldsymbol{\alpha}$ as input to connect the expression motion to the specific person, and the emotion embedding $\boldsymbol{z}_{emo}$ produced from the fixed CLIP text encoder as the emotion condition. Besides, instead of utilizing a one-hot coding~\cite{wang2020mead} to control the emotion intensity, we further prepend a learnable intensity token $\boldsymbol{\varphi}$, which is the product of the base learnable intensity vector and intensity scalar:
\begin{equation}
  \begin{aligned}
    \boldsymbol{\varphi} &= \eta \boldsymbol{\varphi}_{base},
  \end{aligned}
\end{equation}
where $\eta \in \left\{1,2,3\right\}$ at the training phase, and it can be a continuous random value range from 1 to 3 during the testing phase. Typically, the audio sequence and identity token are first embedded into the hidden dimension, then together with the prefix token $\boldsymbol{\varphi}$ to be added with standard positional PE and emotion embeddings:
\begin{equation}
    \begin{aligned}
        &\bar{\boldsymbol{\beta}} = \boldsymbol{\mathcal{T}}([\boldsymbol{\varphi},\text{MLP}(\boldsymbol{\alpha}),\text{MLP}(\boldsymbol{A})] + \text{PE} + \text{MLP}(\boldsymbol{z}_{emo})),
    \end{aligned}
\end{equation}
\begin{equation}
    \begin{aligned}
        \hat{\boldsymbol{\beta}} = \text{MLP}(\bar{\boldsymbol{\beta}}).
    \end{aligned}
\end{equation}

We train EmoA2E independently by two losses. First, we define an expression \textit{Reconstruction Loss} $\mathcal{L}_{rec}$ to calculate the distance between the predicted $\boldsymbol{\hat{\beta}}$ and ground truth $\boldsymbol{\beta}$:
\begin{equation}
  \begin{aligned}
    \mathcal{L}_{rec}= \left\|\boldsymbol{\beta}-\hat{\boldsymbol{\beta}}\right\|_{2}.
  \end{aligned}
\end{equation}
Besides, we further select 68 points from the original 3DMM $\boldsymbol{\rho}$ and modified one $\boldsymbol{\hat{\rho}}$, obtaining $\boldsymbol{l}$ and $\boldsymbol{\hat{l}}$, respectively. We define \textit{Landmark Loss} to measure the similarity between them:
\begin{equation}
  \begin{aligned}
    \mathcal{L}_{lm} = \left\|\boldsymbol{l}-\boldsymbol{\hat{l}}\right\|_{2}.
  \end{aligned}
\end{equation}
Thus, the total loss is defined as follows:
\begin{equation}
  \begin{aligned}
    \mathcal{L} = \lambda_{rec}\mathcal{L}_{rec} + \lambda_{lm}\mathcal{L}_{lm},
  \end{aligned}
\end{equation}
where $\lambda_{rec}=100$ and $\lambda_{lm}=0.1$.

\subsection{Texture-Geometry-aware Diffusion Model}
\label{sec:TGDM}
Most recent GAN-based methods are source-oriented that explicitly model the deformation to animate the source into the driving pose and expression. However, it is still quite challenging to achieve the accurate desired geometry and capture the complex identity appearance when under various extreme conditions, such as large pose, yielding noticeable artifacts and degradation problems. Thus, we revisit this task and propose the target-oriented Texture-Geometry-aware Diffusion Model (TGDM), which focuses on transferring the source texture to the rendered geometry face and inherits the flexibility and fidelity of diffusion models. In this part, we give the descriptions of the network structure and the training details for the denoising process.

\noindent\textbf{Architecture.} Following the~\cite{ho2020denoising}, our conditional denoising model $\boldsymbol{\epsilon}_{\theta}$ is designed by the UNet-based backbone, consisting of the encoder $\boldsymbol{\Psi}_E$ and decoder $\boldsymbol{\Psi}_D$. As shown in Fig.~\ref{fig:pipeline_reenact}, TGDM is conditioned on three external inputs. First, the texture encoder $\boldsymbol{\Phi}_E$ provides the multiscale features $\boldsymbol{F}_s = \left\{\boldsymbol{F}_s^0, \dots,\boldsymbol{F}_s^k\right\}$  to provide the desired texture patterns, where $k$ is 1, \ie, we adopt two resolution texture features in $16 \times 16$ and $32 \times 32$. To mix the source texture within the noise prediction branch and eliminate the effects of misalignment, we design the Texture Attention-based (TexAtt) module that employs the cross-attention mechanism for better integration. As shown in Fig.~\ref{fig:pipeline_reenact}, each TexAtt receives the source texture feature $\boldsymbol{F}_s^i$ and the noise feature $\boldsymbol{F}_d^i$, the query is extracted by one convolution from $\boldsymbol{F}_d^i$, and the key and value are extracted from $\boldsymbol{F}_s^i$ in the same way, obtaining $\boldsymbol{Q}_d, \boldsymbol{K}_s, \boldsymbol{V}_s \in \mathbb{R}^{C_i/4 \times H_i \times W_i}$ with reduced channel numbers. Then $\boldsymbol{Q}_d$ and $\boldsymbol{K}_s$ are used to calculate the correlation matrix $\boldsymbol{M}$, which further multiplies $\boldsymbol{V}_s$ to obtain $\boldsymbol{F}_{s \rightarrow d}^{i}$. A zero-initialized learned scale parameter $\tau$ is applied on $\boldsymbol{F}_{s \rightarrow d}^{i}$ to control the source texture transfer flow when added to the $\boldsymbol{F}_d^i$:
\begin{equation}
    \begin{aligned}
    \boldsymbol{F}_{s \rightarrow d}^i = \text{softmax}(\boldsymbol{Q}_d(\boldsymbol{K}_s)^T)\boldsymbol{V}_s=\boldsymbol{M}\boldsymbol{V}_s,
    \end{aligned}
\end{equation}
\begin{equation}
    \begin{aligned}
    \hat{\boldsymbol{F}_d^i} = \tau \boldsymbol{F}_{s \rightarrow d} + \boldsymbol{F}_d^i.
    \end{aligned}
\end{equation}
Then, the spatially aligned rendered face $\boldsymbol{I}_{3d}$ is concatenated channel-wise with the noisy face $\boldsymbol{Z}_T$, which is obtained by adding noise to $\boldsymbol{I}_d$ according to Eq.~\ref{eq:addnoise}. They are fed to the first layer of the network to guide the denoising process, ensuring the intermediate noise and the output face follow the given facial geometry. In addition, the modified coefficients $\boldsymbol{\hat{\rho}}$ further supplement the implicit geometry cues, especially the gaze direction not included in the rendered face. It added with embedded time, forming the last condition $\boldsymbol{C} = \text{Linear}(\boldsymbol{\hat{\rho}}) + \text{Linear}(t)$, which is injected into the noise predictor via the adaptive instance normalization (AdaIN)~\cite{huang2017arbitrary}:
\begin{equation}
    \begin{aligned}
    \text{AdaIN}(\boldsymbol{F}_d^i, \boldsymbol{C}) = \sigma_c(\boldsymbol{C}) \frac{\boldsymbol{F}_d^i-\mu(\boldsymbol{F}_d^i)}{\sigma(\boldsymbol{F}_d^i)} + \mu_c(\boldsymbol{C}),
    \end{aligned}
\end{equation}
where $\mu(\cdot)$ and $\sigma(\cdot)$ is the average and variance operation of the input feature $\boldsymbol{F}_d^i$ respectively. $\mu_c(\cdot)$ and $\sigma_c(\cdot)$ are used to estimate the adapted mean and bias according to the given condition. To this end, all condition information is properly integrated into the network $\boldsymbol{\epsilon}_{\theta}(\boldsymbol{Z}_t, \boldsymbol{F}_s, \boldsymbol{I}_{3d}, \boldsymbol{\hat{\rho}}, t)$ to predict the noise for talking face generation.

\noindent\textbf{Objectives.} We first adopt the regular simple \textit{Denoising Loss}:
\begin{equation}
    \begin{aligned}
    \mathcal{L}_{simple} = \left\|\boldsymbol{\epsilon}-\boldsymbol{\epsilon}_{\theta}(\boldsymbol{Z}_t, \boldsymbol{F}_s, \boldsymbol{I}_{3d}, \boldsymbol{\hat{\rho}}, t)\right\|_{2},
    \end{aligned}
\end{equation}
where $\boldsymbol{\epsilon}$ is an added noise on $\boldsymbol{I}_d$. Besides, we estimate the fully denoised face $\hat{\boldsymbol{Z}}_0$ according to the Eq.~\ref{eq:noised}, which enables further constraints on the image level. Concretely, we measure the difference between $\boldsymbol{\hat{\boldsymbol{Z}}_0}$ and $\boldsymbol{I}_d$ at the pixel and perceptual level by a \textit{Reconstruction Loss} $\mathcal{L}_{rec}$ as $\mathcal{L}_2$ distance and a \textit{Perceptual Loss} as the LPIPS loss~\cite{zhang2018unreasonable}:
\begin{equation}
  \begin{aligned}
    \mathcal{L}_{rec} = \left\|\hat{\boldsymbol{Z}}_0-\boldsymbol{I}_d\right\|_{2},
  \end{aligned}
\end{equation}
\begin{equation}
  \begin{aligned}
    \mathcal{L}_{p} = \left\|\phi_{vgg}(\hat{\boldsymbol{Z}}_0)-\phi_{vgg}(\boldsymbol{I}_d)\right\|_{2},
  \end{aligned}
\end{equation}
where $\phi_{vgg}(\cdot)$ represents the pre-trained VGG16~\cite{simonyan2014very} network. Thus, the total loss is defined as follows:
\begin{equation}
  \begin{aligned}
    \mathcal{L} = \lambda_{simple}\mathcal{L}_{simple} + \lambda_{rec}\mathcal{L}_{rec} + \lambda_{p}\mathcal{L}_{p},
  \end{aligned}
  \label{eq:3}
\end{equation}
where $\lambda_{simple}=10$, $\lambda_{rec}=1$, and $\lambda_{p}=1$.

\subsection{A Novel Face Swapping Paradigm Built on TGDM}
\label{sec:swap}
Despite the impressive progress of recent methods, GAN- and diffusion-based face swapping methods still suffer from the dilemma that the improvement of source face identity consistency at the expense of sacrificing target attribute preservation. For example, DiffFace~\cite{diffface} employs identity and attribute expert models to guide the noise prediction, and the balance between them is critical to producing high-quality swapped faces. However, it is complex and needs many experimental attempts. We attribute this phenomenon to the training phase playing the seesaw-style game, which struggles to balance all identity-unrelated attributes preservation and the source identity fusion. Since our proposed method for talking face is able to transfer complex textures, we derive a novel paradigm for face swapping built upon the TGDM.

\begin{figure}[t!]
	\centering
	\includegraphics[width=0.48\textwidth]{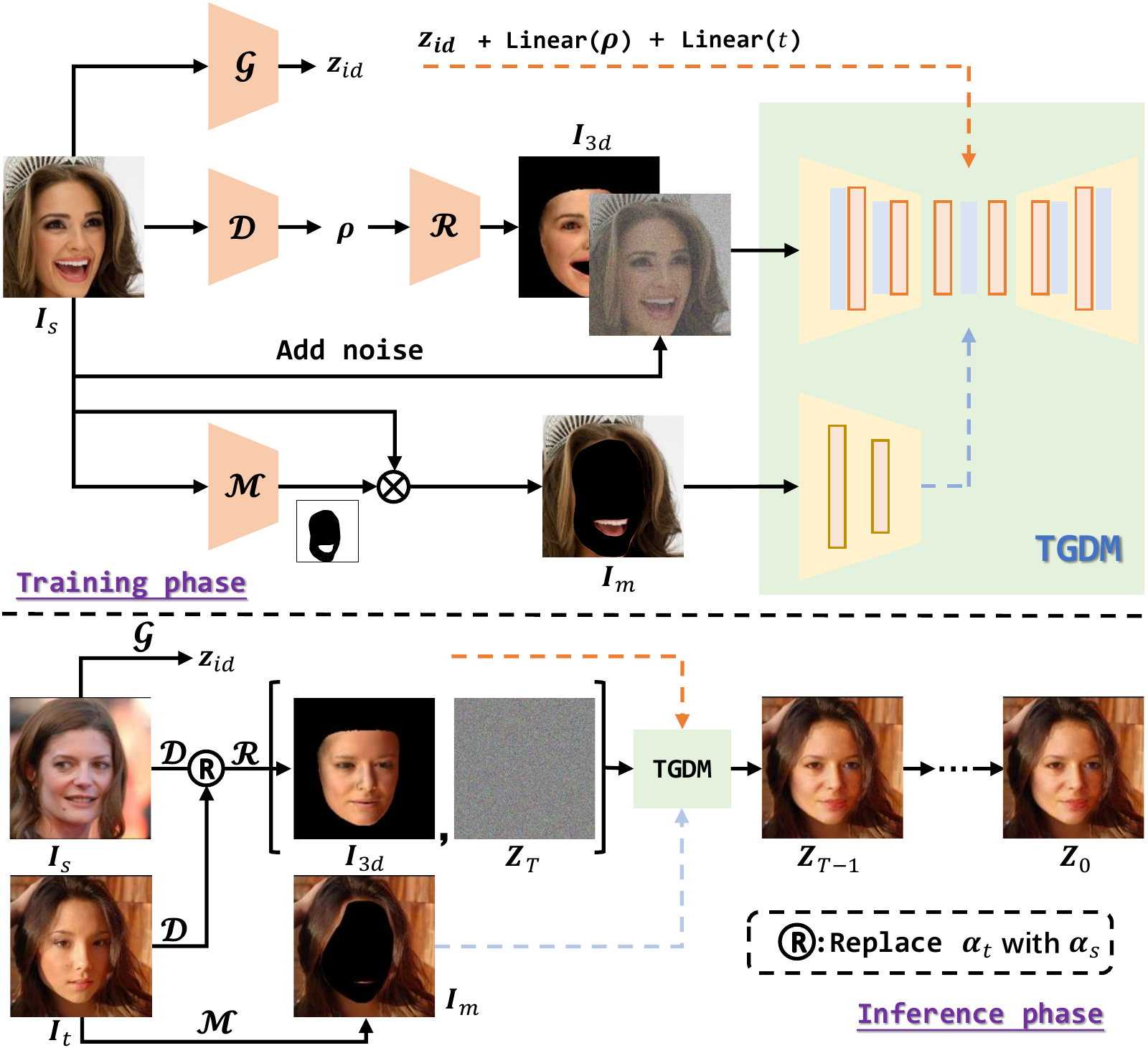}
	\caption{The new paradigm of face swapping is built upon the TGDM. During the training phase, we focus on reconstructing the input face $\boldsymbol{I}_s$ from given identity- and attribute-related conditions, \ie, identity embedding $\boldsymbol{z}_{id}$, 3D face descriptor $\boldsymbol{\rho}$, rendered face $\boldsymbol{I}_{3d}$, and masked face $\boldsymbol{I}_m$. During inference, we first render the recombined coefficients to capture the desired geometry prior, which incorporates with other corresponding conditions for generating final swapped results.
	}
	\label{fig:pipeline_swap}
\end{figure}

\begin{figure*}[t!]
	\centering
	\includegraphics[width=0.85\textwidth]{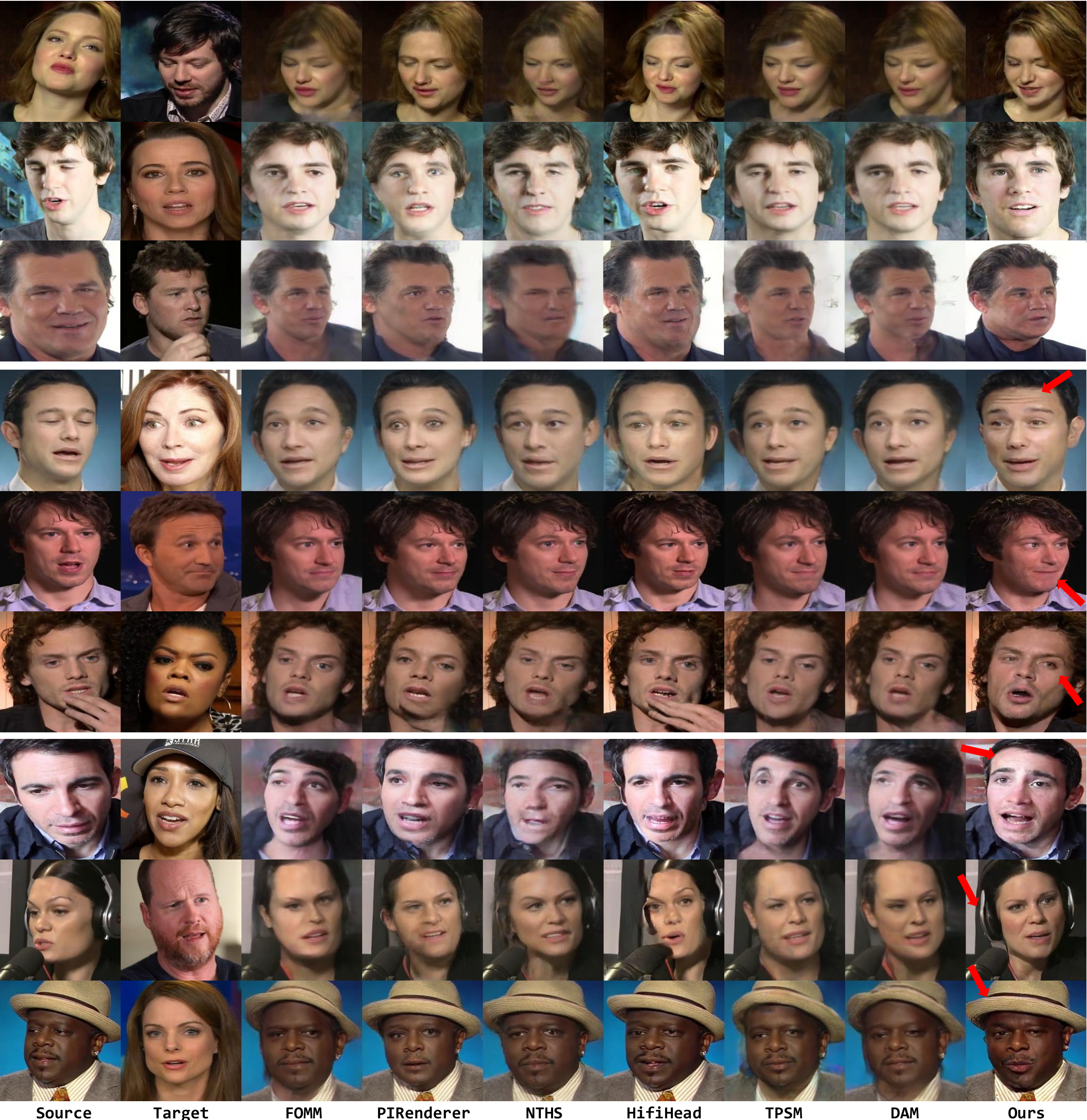}
	\caption{Qualitative comparison with SOTA methods on VoxCeleb1 test set. We present various challenging cases which show the significant difference between the source and target of the pose, occlusion, and face size. Please pay attention to the area indicated by the red arrow.}
	\label{fig:exp_reenactment}
\end{figure*}

\begin{table*}[t]
   \centering
   \scriptsize
   \renewcommand\arraystretch{1.2}
   \setlength\tabcolsep{4pt}
   \begin{tabular}{C{25pt}C{25pt}C{25pt}C{25pt}C{25pt}C{25pt}C{25pt}C{25pt}C{25pt}C{25pt}C{25pt}C{25pt}C{25pt}}
      \toprule
      \multicolumn{1}{c}{\multirow{2}{*}{Method}} & \multicolumn{7}{c}{\textbf{Same-Identity}} & \multicolumn{5}{c}{\textbf{Cross-Identity}}  \\
      \cmidrule(lr){2-8} \cmidrule(lr){9-13}
      & PSNR $\uparrow$  & LPIPS $\downarrow$ & Exp $\downarrow$ & Angle $\downarrow$ & Gaze $\downarrow$ & ID-C $\uparrow$ & FID $\downarrow$ & Exp $\downarrow$ & Angle $\downarrow$ & Gaze $\downarrow$ & ID-C $\uparrow$ & FID $\downarrow$ \\
      \midrule
      FOMM  & 16.32 & 0.3459 & 5.52 & 0.0474 & 0.0749 & 0.6552 & 27.56 & 7.14 & 0.0613 & 0.0961 & 0.5445 & 41.77\\
      PIRenderer  & 16.72 & 0.3549  & 5.41 & 0.0546 & 0.0773 & 0.6576 & 28.90 & 6.90 & 0.0673 & 0.0971 & 0.5503 & 37.95 \\
      NTHS  & 18.13 & 0.3588 & 5.98 & 0.0625 & 0.0903 & 0.7091 & 27.07 & 7.77 & 0.0814 & 0.1166 & 0.6159 & 38.48 \\
      HifiHead  & 15.72 & 0.3678 & 5.39 & 0.0693 & \underline{0.0625} & \textbf{0.8722} & \textbf{21.53} & \underline{6.80} & 0.0871 & \underline{0.0746} & \textbf{0.8394} & \textbf{33.77} \\
      TPSM  & \textbf{19.29} & \underline{0.3366} & \underline{5.28} & \underline{0.0412} & 0.0660 & 0.6918 & 25.57 & 6.88 & \underline{0.0536} & 0.0853 & 0.5917 & 39.28 \\
      DAM  & 18.05 & 0.3440 & 5.46 & 0.0484 & 0.0737 & 0.6535 & 28.11 & 7.08 & 0.0626 & 0.0949 & 0.5415 & 44.09 \\
      Ours  & \underline{18.55} & \textbf{0.3346} & \textbf{5.09} & \textbf{0.0315} & \textbf{0.0554} & \underline{0.7718} & \underline{25.51} & \textbf{5.82} & \textbf{0.0349} & \textbf{0.0596} & \underline{0.7017} & \underline{35.16} \\
      \bottomrule
   \end{tabular}
   \caption{Quantitative results on the tasks of same-identity and cross-identity setting on VoxCeleb1. Bold and underline represent optimal and suboptimal results. The up arrow indicates
   that the larger the value, the better the model performance, and vice versa.}
   \label{tab:sota_reenactment}
\end{table*}

\begin{figure}[t!]
	\centering
	\includegraphics[width=0.48\textwidth]{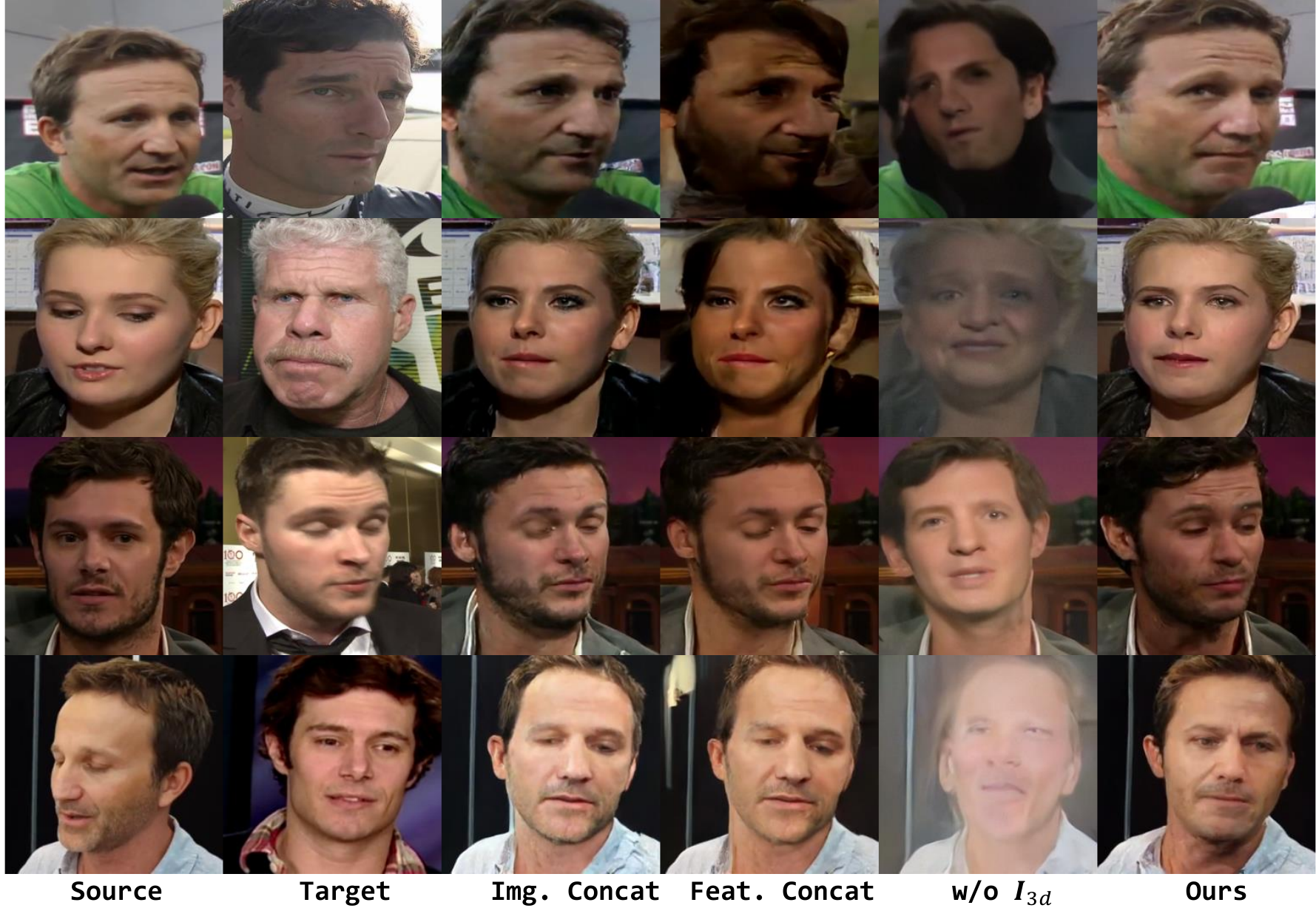}
	\caption{Qualitative ablation study of our method with different variations on VoxCeleb1 dataset.}
	\label{fig:exp_reenactment_abla}
\end{figure}

\begin{figure}[t!]
	\centering
	\includegraphics[width=0.31\textwidth]{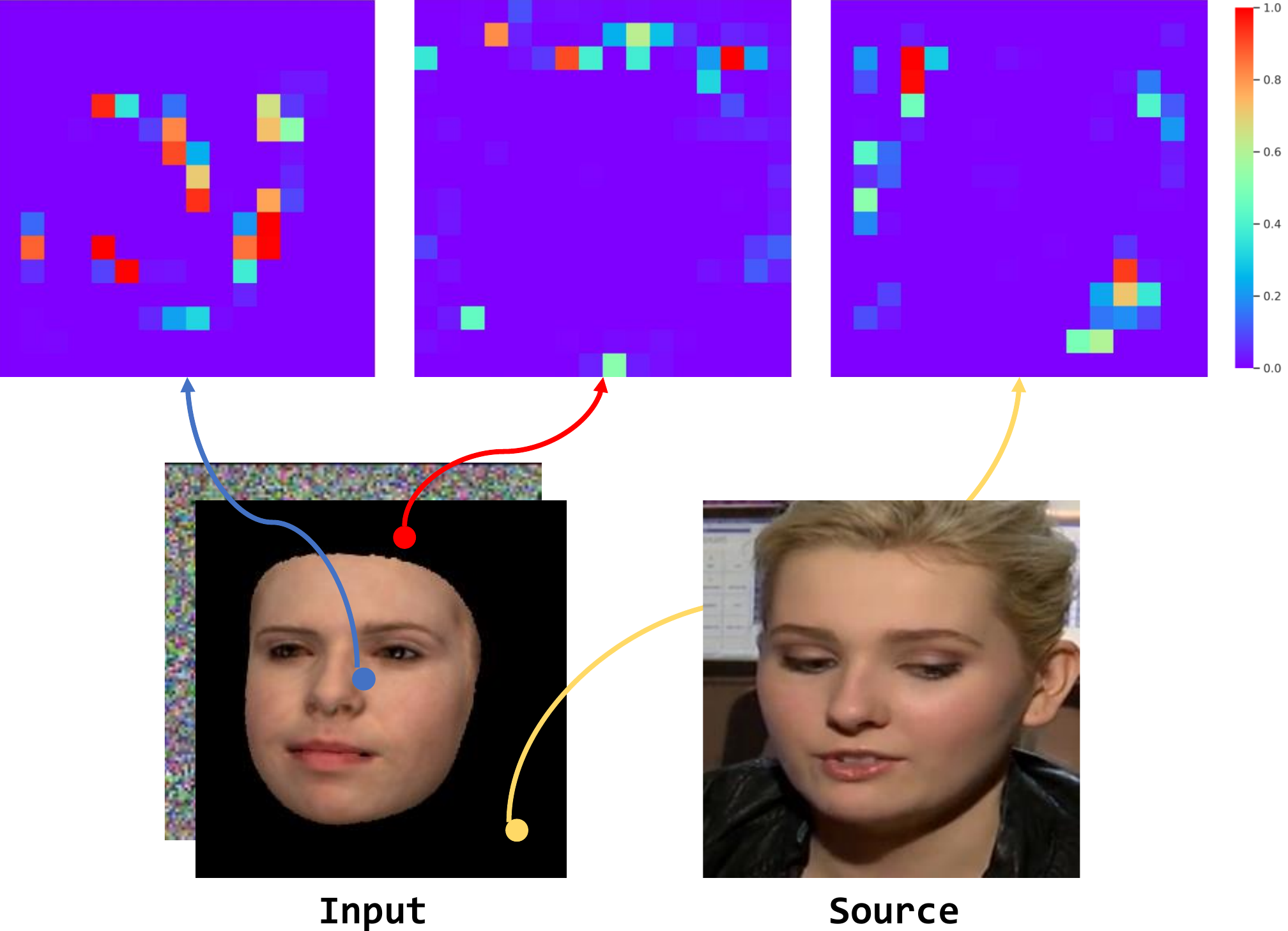}
	\caption{Attention visualization of TextAtt. The color bars indicate activation values. The points in the input rendered face could correctly match similar semantic and geometrical areas in the source.}
	\label{fig:att}
\end{figure}

Specifically, as shown in the top of Fig.~\ref{fig:pipeline_swap}, there are two modifications. First, we completely mask the face region of the source texture image with the help of the mask predictor $\mathcal{M}$~\cite{yu2018bisenet} to ensure that the ground truth identity information is not visible to the network. Second, because of the low-dimensional linear representation of 3DMMs, the rendered images often lack photo-realism and fine texture details like wrinkles. We further supplement the identity embedding from the expert identity model $\mathcal{G}$~\cite{huang2020curricularface}. In this way, the renderer image $\boldsymbol{I}_{3d}$, identity embedding $\boldsymbol{z}_{id}$, and $\text{Linear}(\boldsymbol{\rho})$ focused on affording identity cues and identity-unrelated attributes of the face region, while $\boldsymbol{I}_m$ makes up for the absence of hair and background. Notably, the mouth area is also served as the background, which is discussed in the Sec.~\ref{sec:5.4}. During training, as Eq.~\ref{eq:3}, our scheme does not require complex losses. Instead, the reconstruction loss is sufficient. The hyperparameter setting is the same as Eq.~\ref{eq:3} either.
For inference, given the source $\boldsymbol{I}_s$ and the target $\boldsymbol{I}_t$, we first render the $\boldsymbol{I}_{3d}$ with the identity factor of the source and the remaining parameters of the target. As shown in the bottom of Fig.~\ref{fig:pipeline_swap}, $\boldsymbol{I}_{3d}$ is sensitive to the geometric structure, exhibiting the exact desired face shape, and $\boldsymbol{z}_{id}$ contains source identity semantics. Combining both of them guarantees identity similarity. To this end, following the standard denoising process, our method successfully transfers the source geometry- and semantic-aware identity information to the target, while fully keeping the identity-unrelated attributes without any complex sampling tricks.

\section{Experiment}
\subsection{Datasets and Implementation Details}
\noindent\textbf{Datasets.} For talking face generation, we leverage the VoxCeleb1~\cite{nagrani2017voxceleb} dataset, which contains over 20K videos. Among them, We select the high-resolution (720P) ones and follow the preprocessing method in FOMM~\cite{siarohin2019first} to crop the videos and resize them to $256 \times 256$, obtaining 17,927 training videos and 491 testing videos. For emotional talking face generation, we adopt the MEAD~\cite{wang2020mead} dataset, which contains eight emotion types (neutral, angry, contempt, disgusted, fear, happy, sad, and surprised) and three intensity levels (levels 1, 2, 3). We randomly select 36 identities of front-view video clips for training and the rest for testing. For face swapping, we utilize the high-quality CelebAMask-HQ~\cite{celebAMask-HQ} dataset, which has 30,000 images with fine-grained mask annotation. FaceForensics++~\cite{rossler2019faceforensics++} is used for testing, which is a forensics dataset consisting of 1000 videos.

\noindent\textbf{Metrics.} For face reenactment, we use PSNR and LPIPS~\cite{zhang2018unreasonable} to evaluate reconstruction quality. Exp, Angle, and Gaze are used to calculate the average Euclidean Distances of corresponding coefficients between the generated and target faces. We employ ID embeddings extracted by Curricularface~\cite{huang2020curricularface} (ID-C) and Arcface~\cite{deng2019arcface} (ID-A) to measure identity cosine similarity. We further use FID~\cite{heusel2017gans} to evaluate the realism of the generated faces. For talking face generation, in addition to the metrics mentioned above, we use Landmarks Distance (LMD)~\cite{chen2018lip} around the mouth, and the confidence score (Sync) proposed in SyncNet~\cite{chung2017out} to measure the accuracy of mouth shapes and lip synchronization. We further use Emotion Feature Distance (EFD) to measure the accuracy of the emotion representation, which is extracted by~\cite{mollahosseini2017affectnet}. For face swapping, we adopt Exp, Angle, ID-A, and FID for evaluation. We do not adopt ID-C since Curricularface has been used in training and inference.

\noindent\textbf{Implementation Details.} For EmoA2E, we randomly sample consecutive $K=32$ clips for emotion-condition training in MEAD and emotion-free training in VoxCeleb1. We use a learning rate of 0.0002 and 128 batch sizes with the Adam optimizer on one V100 GPU for 200K iterations. For TGDM, we randomly sample the source and target faces from the same video in MEAD and VoxCeleb1 for training. It takes about 4 days by using 4 V100 GPUs with 8 batch sizes and a 0.0002 learning rate for 200K iterations. For face swapping, we train its model as the aforementioned setting for approximately 3 days. For the diffusion model, the length of the denoising step $T$ is set to 1000, and a linear noise schedule is adopted for both the training and inference process. Notably, to stale the training procedure, only MSE loss of noise is used at the beginning of the training. Only when it has been decreased below 0.05, MSE loss of image, and LIPIS loss then start to work. Besides, the UNet of TGDM receives $256 \times 256$ resolution images and performs 16 down-sample ratios.

\begin{table}[t]
   \centering
   \scriptsize
   \renewcommand\arraystretch{1.2}
   \setlength\tabcolsep{4pt}
   \begin{tabular}{C{35pt}C{25pt}C{25pt}C{25pt}C{25pt}C{25pt}}
      \toprule
      Method & Exp $\downarrow$ & Angle $\downarrow$ & Gaze $\downarrow$ & ID-C $\uparrow$ & FID $\downarrow$ \\
      \midrule
      Img Concat & \underline{6.06}  & \underline{0.0353}   & \underline{0.1252}   & \underline{0.6323}	            & \underline{46.63}                  \\
      Feat Concat & 6.69  & 0.0443   & 0.1346   & 0.5204	            &  52.96                \\
      w/o $\boldsymbol{I}_{3d}$ & 9.10  & 0.4376   & 0.1845   & 0.2233	            &    70.49               \\
      Ours & \textbf{5.82}  & \textbf{0.0349}   & \textbf{0.0596}   & \textbf{0.7017}	            & \textbf{35.16}                  \\
      \bottomrule
   \end{tabular}
   \caption{Quantitative ablation study of our approach with different module on VoxCeleb1.}
   \label{tab:abla_reenactment}
\end{table}

\begin{figure*}[t!]
	\centering
	\includegraphics[width=0.82\textwidth]{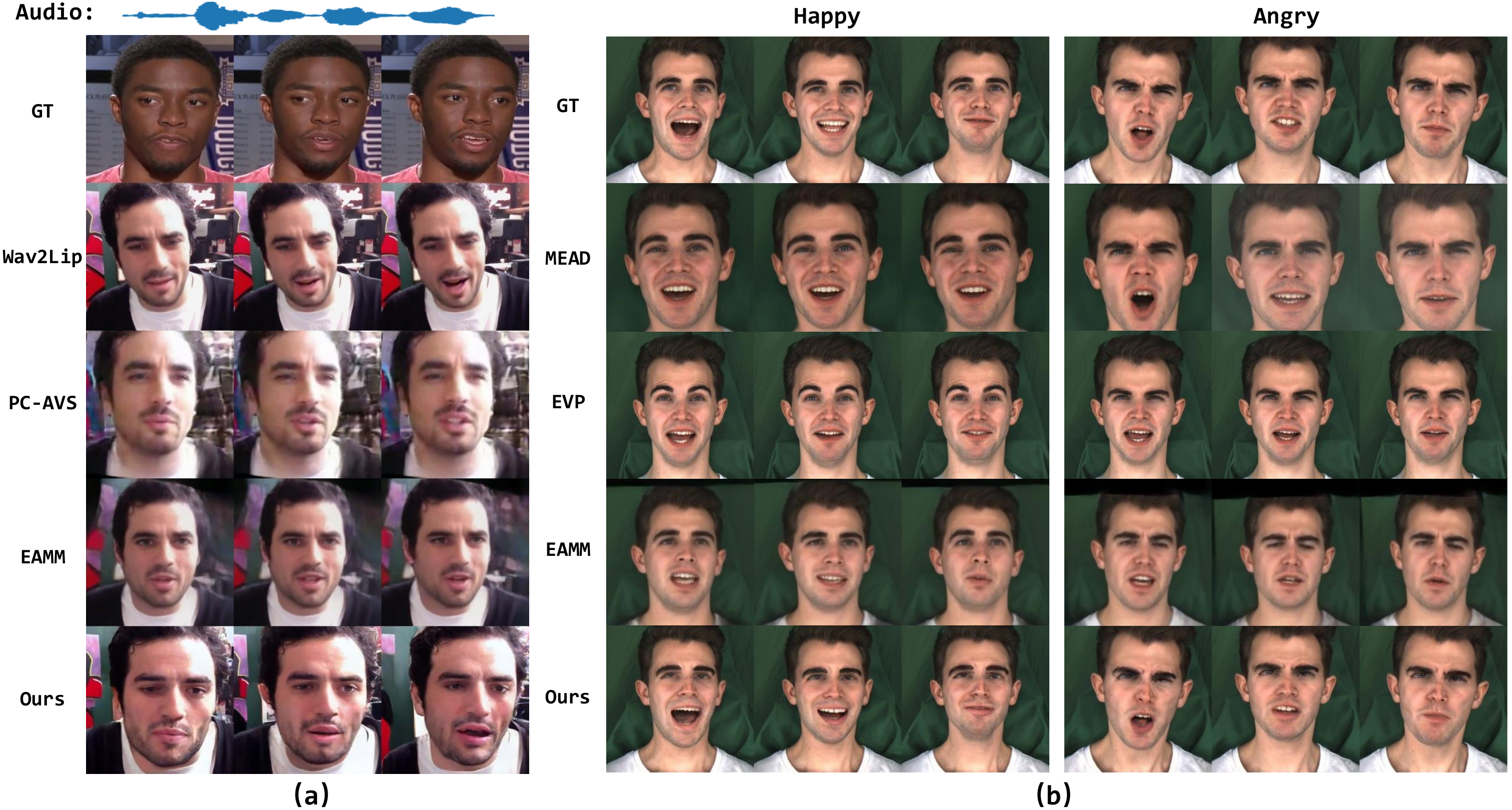}
	\caption{(a) presents qualitative results of audio-driven talking face generation in VoxCeleb1 under the cross-identity setting, while (b) takes emotion into consideration. The results are sampled from the MEAD dataset. Different columns mean several sampled timestamps. Images are from officially released codes for fair comparisons.}
	\label{fig:emo_sota}
\end{figure*}

\subsection{Face Reenactment}
\label{sec:5.2}
\subsubsection{Comparison with Baselines}
\noindent\textbf{Qualitative Results.} We perform qualitative comparisons with FOMM~\cite{siarohin2019first}, PIRenderer~\cite{ren2021pirenderer}, NTHS~\cite{wang2021one}, HifiHead, TPSM~\cite{zhao2022thin}, and DAM~\cite{tao2022structure} in the \textit{Cross-Identity} setting, where the source and the target are of different identities. We do not compare with StyleHeat~\cite{yin2022styleheat} since it requires the aligned inputs due to the fixed StyleGAN generator. As shown in Fig.~\ref{fig:exp_reenactment}, we sample nine pairs from VoxCeleb1 for visualization. First, the top three pairs have a significant difference in face size. It can be seen that FOMM-based methods, \eg, TPSM and DAM, produce over-smooth facial textures and suffer from noticeable warping artifacts. HifiHead could generate realistic faces, but their poses are inconsistent with the target. By contrast, the results of our method are of high quality and with the desired attributes. 
Second, the target faces of the middle ones show rich micro-expressions. Recent methods just imitate mouth shape and head direction, and they ignore the emotion embodied in the target. For example, the target of the fourth row is surprised, and the sixth is contempt. For comparison, our results exhibit accurate emotion styles, \ie, surprised forehead lines, delighted mouth corners, and disdainful eyes. Finally, the bottom pairs suffer from occlusions in the source or the target. It is difficult for FOMM-based methods to estimate the precise key points, thus usually resulting in extremely distorted facial shapes (the head area of row 7). Other methods also struggle to animate the occluded objects to fit the desired pose. Benefiting from the effective cross-attention mechanism, our method is not sensitive to occlusion and reasonably preserves the non-facial parts in the generated results (the headphones of row 8 and the hat of row 9). Moreover, these cases are all under large-pose conditions, which convincingly demonstrate that our method successfully transfers the source texture to the target rendered image, providing more realistic results with accurate pose and detailed expression while preserving the source identity.

\noindent\textbf{Quantitative Results.} We quantitatively compare the proposed method with several aforementioned SOTA methods both in \textit{Same-Identity} and \textit{Cross-Identity} settings. We randomly sample 200 identities from the test set and set 5 random seeds to generate 1K pairs in total. The results are summarized in Tab.~\ref{tab:sota_reenactment}. Benefiting from the explicit facial representation contained in the rendered face, our methods achieve an impressive performance of facial attributes, indicating that our model can animate the source face that is highly faithful to the given structure cues. Furthermore, our method is favorable against other methods regarding the reconstruction metrics PSNR and LPIPS and image quality metric FID. HifiHead obtains the lowest FID due to the StyleGAN-based generator. Besides, it also shows the best identity consistent but suffers from severe pose error, which can be concluded from rows 2 and 8 of Fig~\ref{fig:exp_reenactment} either. Overall, the above observations are consistent with the qualitative results in Fig.~\ref{fig:exp_reenactment}.

\subsubsection{Ablation Study and Analysis}

\noindent\textbf{Ablation Study.} We perform qualitative and quantitative experiments to validate the
merits of the proposed designs. Specifically, we design two variations to evaluate the effectiveness of TextAtt. Specifically, we adopt the image-level and feature-level concatenation for feature injection as two baselines. For a fair comparison, we train our method and two baselines with the same setting, \eg, same batch sizes and training iterations. As shown in columns 3 and 4 of Fig.~\ref{fig:exp_reenactment_abla}, these two baselines are able to generate the desired pose and expression, but they have a limited ability to retain the source appearance, exhibiting severe color jitting, especially the feature-level concatenation. To further verify the necessity of the rendered face $\boldsymbol{I}_{3d}$, we only use the 3D face descriptors $\boldsymbol{\rho}$ to supply the facial geometry information. Comparing the results of columns 5 and 6, we can observe that the implicit representation is insufficient to effectively support geometry alignment. Contrary to the above competitors, our results show higher quality, which illustrates the effectiveness of cross attention as the feature transfer module and rendered image as the explicit geometry condition, both reducing the difficulty of training and speeding up the convergence of the model. Besides, the above observations could also be summarized from Tab.~\ref{tab:abla_reenactment}, our proposed method improves all metrics by a large margin.

\noindent\textbf{Interpretability of TextAtt.} To better understand the cross-attention mechanism, we visualize the attention maps of the TextAtt in UNet middle block, which is $16 \times 16$ resolution. As shown in Fig.~\ref{fig:att}, we select three points from different regions in the noise feature, \ie, head, face, and background. The visualized attention maps indicate that each location pays more attention to the geometrically and semantically similar areas, \eg, the red point is sampled from the head region, which has a higher response with the corresponding region of the source feature. Consequently, the such attention-based design allows the explicit texture transfer to achieve photo-realistic and identity-consistent face generation.

\begin{figure}[t!]
	\centering
	\includegraphics[width=0.48\textwidth]{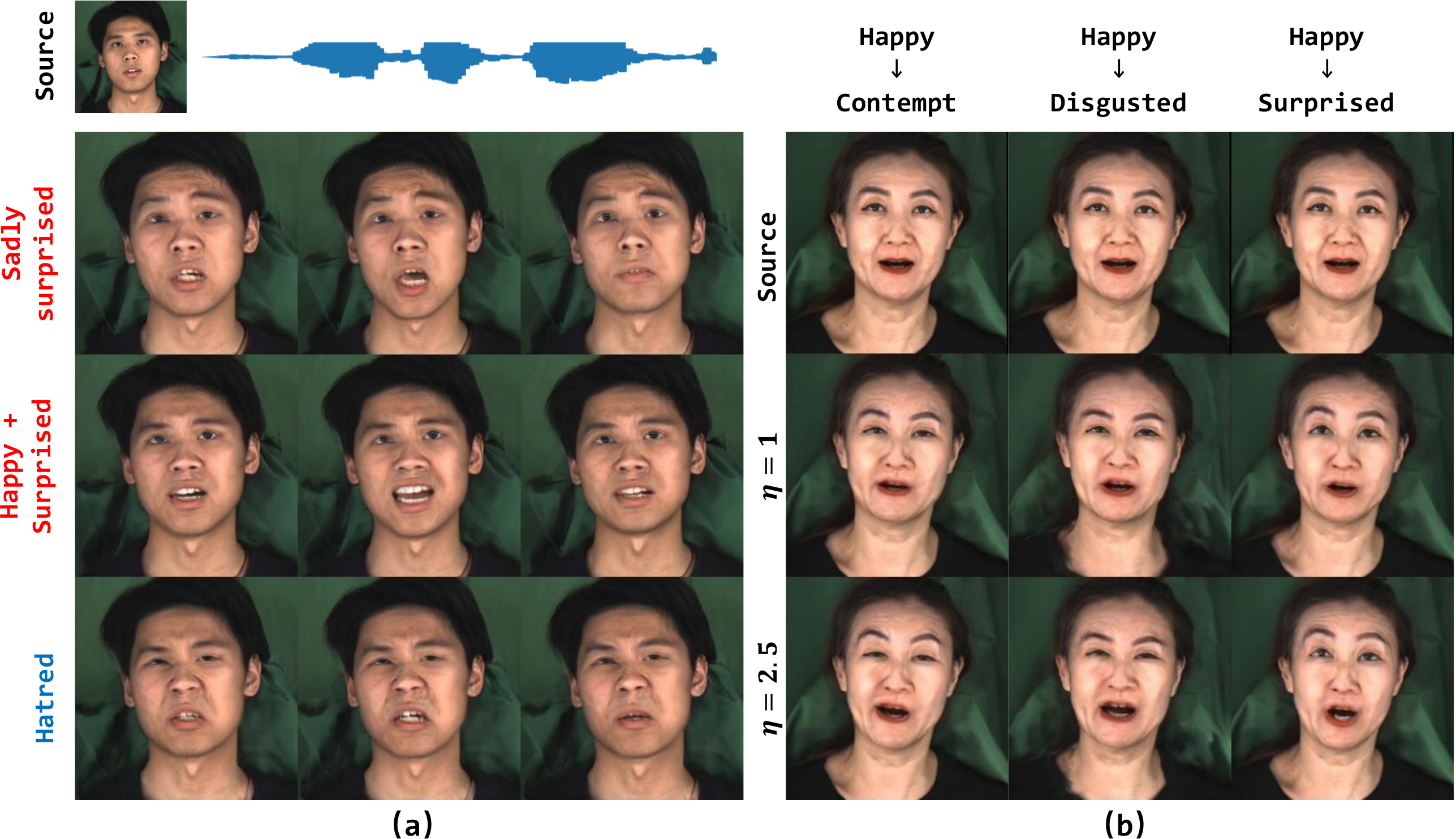}
	\caption{(a) The visualization of unseen emotion style. Rows 2 and 3 (in {\color{red}Red}) are the compound styles, and row 4 (in {\color{blue}Blue}) is a new style. (b) Results of different emotion styles and intensity levels.}
	\label{fig:emo_abla}
\end{figure}

\begin{table}[t]
   \centering
   \scriptsize
   \renewcommand\arraystretch{1.2}
   \setlength\tabcolsep{4pt}
   \begin{tabular}{C{55pt}C{25pt}C{25pt}C{25pt}C{25pt}C{25pt}}
      \toprule
      Method & EFD $\downarrow$ & LMD $\downarrow$ & Sync $\uparrow$ & ID-C $\uparrow$ & FID $\downarrow$ \\
      \midrule
      Wav2Lip & -  & \underline{3.09}   & \underline{4.86}   & -	            &     -                      \\
      PC-AVS & -   & 3.21   & 4.65   & \underline{0.81}	 & \underline{30.72}                          \\
      EAMM-Neutral & - & 3.22 & 4.60 & 0.79 & 37.90 \\
      Ours & - & \textbf{3.09} & \textbf{4.91} & \textbf{0.83}  & \textbf{26.54} \\
      \midrule
      MEAD & \underline{0.084}  & 2.62   & 3.09   & \underline{0.81}	            & 30.69                         \\
      EVP & 0.106   & 2.54   & 3.21   & 0.70	            & \textbf{12.83}                       \\
      EAMM-Emo & 0.092   & \underline{2.50}   & \underline{3.26}   & 0.74	            & 29.01                         \\
      Ours& \textbf{0.061}  & \textbf{2.36}    & \textbf{3.52}    & \textbf{0.81}	   & \underline{16.33}      \\
      \bottomrule
   \end{tabular}
   \caption{The top part is the quantitative comparison of emotion-free on VoxCeleb1 and the bottom part is the emotion-condition on MEAD dataset.}
   \label{tab:emo_sota}
\end{table}

\begin{figure*}[t!]
	\centering
	\includegraphics[width=0.85\textwidth]{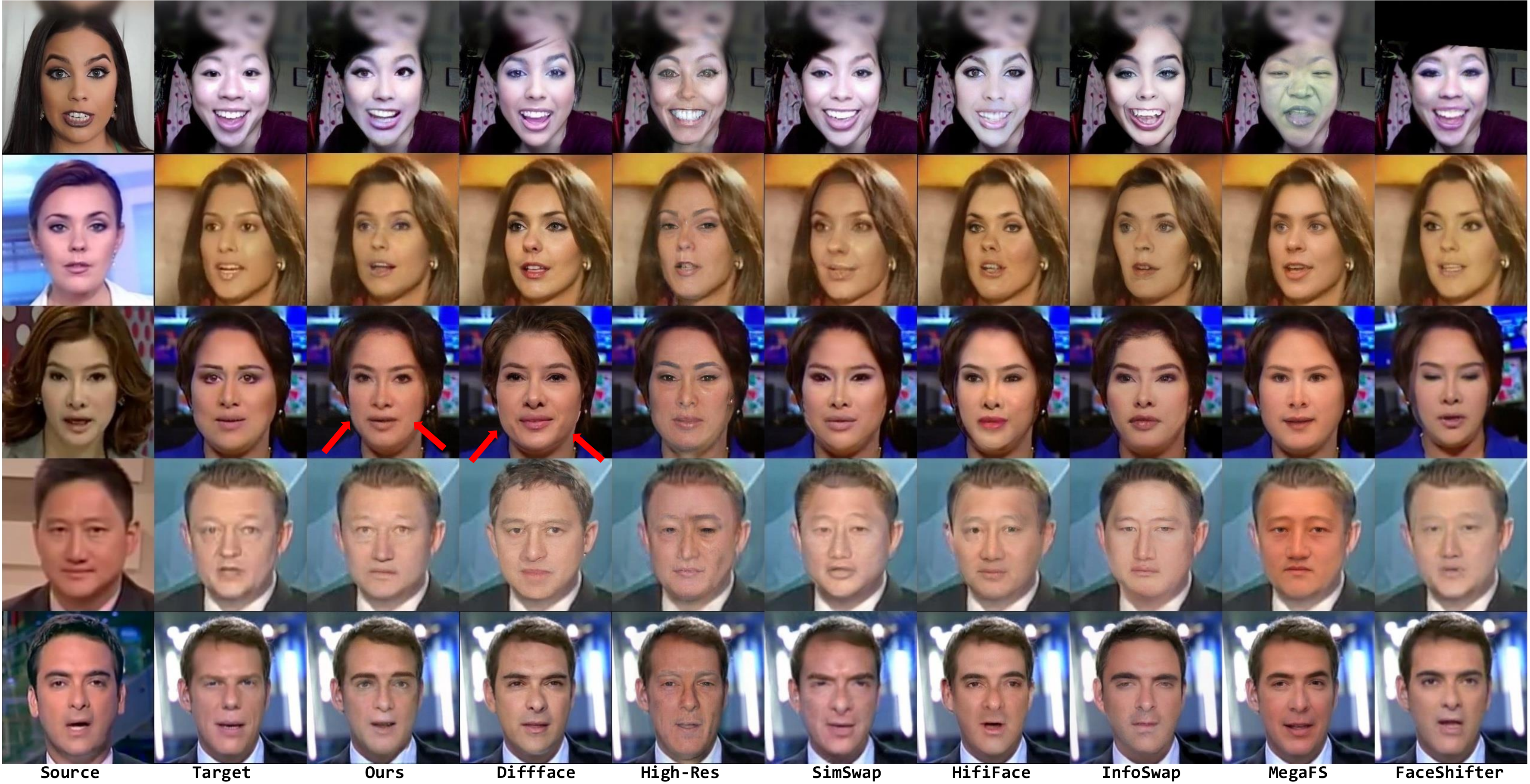}
	\caption{Qualitative comparison of face swapping results with other SOTA models on FaceForensics++. The results of our model better reflect the source identity, especially the face shape. Additionally, they are more faithful to the target image for non-identity-related attributes. The results of other methods are from the DiffFace~\cite{diffface} main paper and officially released codes.}
	\label{fig:swap_sota}
\end{figure*}

\subsection{Talking Face Generation}
\subsubsection{Comparison with Baselines}
\noindent\textbf{Qualitative Results.} We perform qualitative comparisons with Wav2Lip~\cite{prajwal2020lip}, PC-AVS~\cite{zhou2021pose}, and EAMM~\cite{ji2022eamm} for talking face generation. Fig.~\ref{fig:emo_sota} (a) visualizes the generated frames of these methods. It can be seen that all methods could produce synchronized lip shapes with given audio signals. However, Wav2Lip fails to change the head pose and artifacts appear around the mouth area due to the inevitable blending mismatch. PC-AVS and EAMM only support the aligned faces as inputs. Thus such preprocess operation destroys the original facial structure, reflecting on the misalignment pose with the target in the animated faces. Besides, their results are blurred and lose the sharp source textures. By contrast, our results are highly faithful to the given pose from the images and mouth movements from the audio, while maintaining the source texture well. For emotional talking face generation, we select three frames of two emotion styles in MEAD for comparison. As shown in Fig.~\ref{fig:emo_sota} (b), Wav2Lip and PC-AVS struggle to generate desired
emotions with synchronized lip shapes in this task, while the synthesized images from MEAD are of poor quality. EVP~\cite{ji2021audio} and EAMM suffer identity inconsistency with the source and show less rich expression due to lacking intensity modeling. Benefiting from sufficient emotion semantics learning and the powerful generative capabilities of diffusion models, our method produces more accurate expressions and realistic textures.

\noindent\textbf{Quantitative Results.} We conduct a quantitative comparison in the reconstruction setting that guarantees access to ground truth for evaluation. For a fair comparison, we align the cropping manner of all the methods. For talking face generation, as shown in the top part of Tab.~\ref{tab:emo_sota}, we do not calculate the ID-C and FID of Wav2Lip since it only generates the mouth region and copies other regions from input faces. Contrary to other methods, our method yields the best motion control, temporal coherence, identity consistency, and image quality in terms of LMD, Sync, ID-C, and FID. We further compare our emotion-condition pipeline with other emotional talking face generation methods. As shown in the bottom part of the Tab.~\ref{tab:emo_sota}, our method outperforms most metrics except for the FID. EVP achieves higher FID due to the vid2vid-based generator, but it exhibits a weak manipulated ability, which can be inferred from the lower ID-C and Sync, and higher LMD. Moreover, compared with the performance of emotion-free talking face generation and the emotion-condition one, the latter achieves better mouth shape accuracy (lower LMD) due to the limited corpus of MEAD, but the emotions introduce the irregular talking rhythm, leading to the poor synchronization (lower Sync).

\subsubsection{Ablation Study and Analysis}
\noindent\textbf{Ablation Study.} To verify the effectiveness of the Transformer encoder in EmoA2E, we replace it with stacked fully-connected layers of GRU-based recurrent neural networks. Our method outperforms the above two architectures on LMD metric: 3.54 \vs 2.47 \vs \textbf{2.36} of MLPs, GRUs, and Transformers. To further explore the effect of different emotion encoding manners on the unseen emotion style, we use one-hot encoding and language pre-trained model GPT2~\cite{radford2019language} for evaluation. It is obvious that one-hot fails to represent a new style due to the fixed pattern. GPT2 is not available to the visual cues and struggles to reflect the unseen textual semantics to the image domain. We conduct a quantitative experiment that measures the cosine similarity of the attached sequences in Fig.~\ref{fig:emo_abla} with the corresponding text prompts when encoded by GPT2 and CLIP. Our method achieves better results: \textbf{0.621} \vs 0.430, which demonstrates the superiority of CLIP in handling multimodal information.

\noindent\textbf{Generalizing to Unseen Emotion Styles.} Unseen emotion styles include compound and totally new styles. As shown
in Fig.~\ref{fig:emo_abla} (a), row 2 shows the results of the given Sadly surprised, and row 3 of the average embedding of Happy and Surprised, which indicates the flexible manipulation for compound emotion. We further present the new style Hatred in the fourth row. The correct exhibition of these unseen styles verifies the flexibility and rich semantic priors of the CLIP feature space.

\noindent\textbf{Continuous Emotion Style Control.} We conduct a qualitative experiment to evaluate the effectiveness of our method for controlling emotion style. As shown in Fig.~\ref{fig:emo_abla} (b), our approach could change the emotion representation between two distinct styles, rather than previous techniques only taking a neutral face as the source. We increase the intensity value from 1 to 2.5, which shows continuous and accurate expression changes. Please pay attention to the mouth and eyes regions.

\subsection{Expanded Application of Face Swapping}
\label{sec:5.4}
\subsubsection{Comparison with Baselines}
\begin{figure}[t!]
	\centering
	\includegraphics[width=0.45\textwidth]{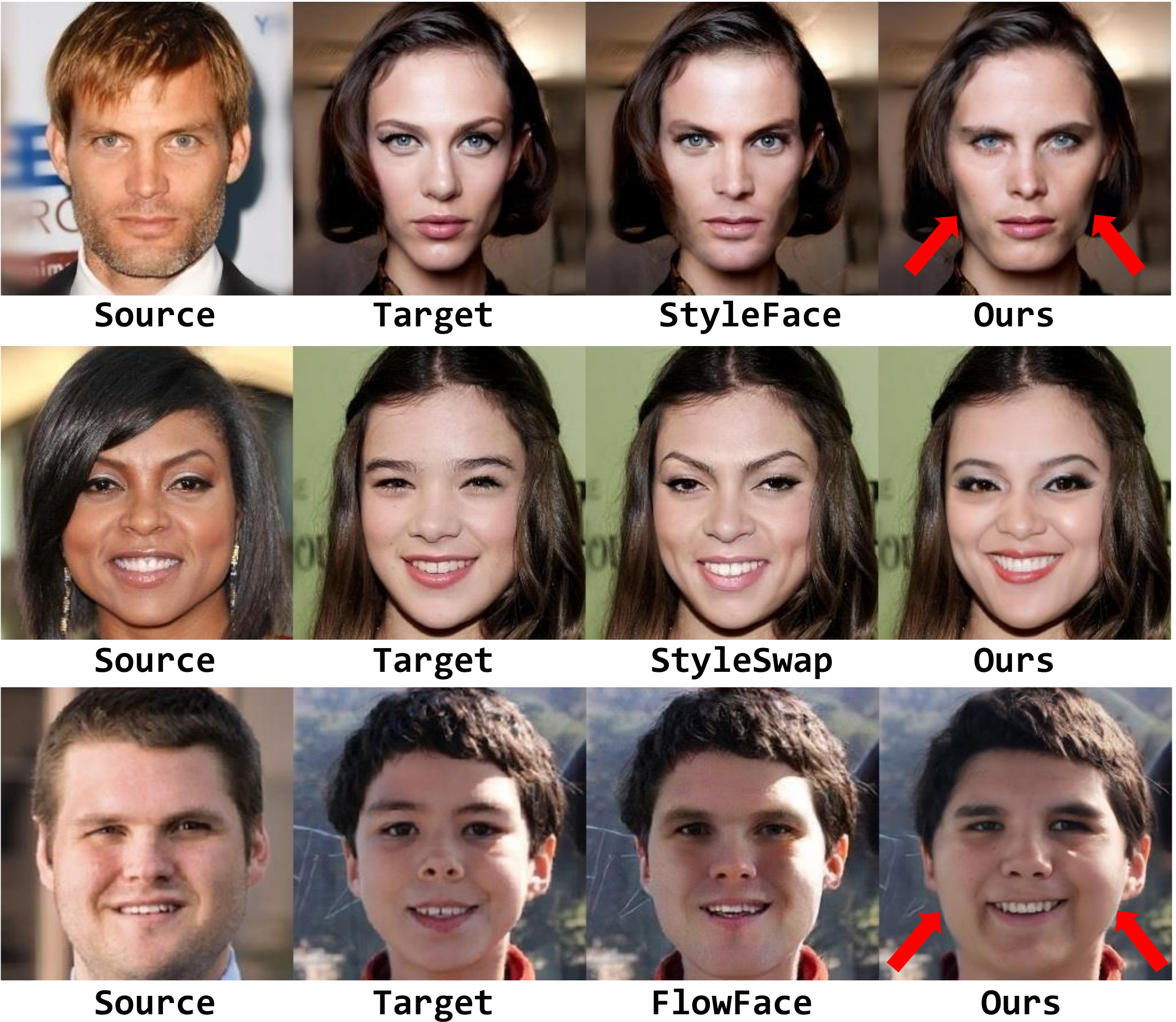}
	\caption{Qualitative comparison of face swapping results with other SOTA models without officially released codes. The inferred images are directly cropped from their papers.}
	\label{fig:swap_sota_2}
\end{figure}

We first conduct qualitative experiments to compare our method with DiffFace~\cite{diffface}, High-Res~\cite{xu2022high}, InfoSwap~\cite{gao2021information}, MegaFS~\cite{wang2021one}, HifiFace~\cite{wang2021hififace}, Simswap~\cite{chen2020simswap}, and FaceShifter~\cite{faceshifter} on the FaceForensics++~\cite{rossler2019faceforensics++} dataset.
As shown in Fig.~\ref{fig:swap_sota}, our model outperforms other models in changing identity-related geometry, especially the face shape, and preserving non-identity-related attributes. For example, in the third row, the generated face shape is more similar to the source, while other methods almost contain the same face shape as the target. Also, in the fourth and fifth rows, we totally preserve the non-identity-related attributes like hair and backgrounds. In the first row, our result is more similar to the source than others. Compared with another diffusion-based method DiffFace, our results obviously show the superiority of generating both identity-consistent and attributes-preserving faces, but the visual quality reduces to some degree. This is because our synthesized faces are more faithful to the target, while DiffFace produces clear but inconsistent textures. Moreover, Fig.~\ref{fig:swap_sota_2} presents more qualitative comparisons with other SOTA methods that without officially released codes, e.g., StyleFace~\cite{styleface}, StyleSwap~\cite{styleswap}, and FlowFace~\cite{zeng2022flowface}. Please attention to the area indicated by the red arrow. We further report quantitative results compared to a part of the above method with officially released codes. The results in Tab.~\ref{tab:swap_sota} also prove that our method is better considering both identity consistency with the source and attribute preservation with the target. 

\begin{figure}[t!]
	\centering
	\includegraphics[width=0.45\textwidth]{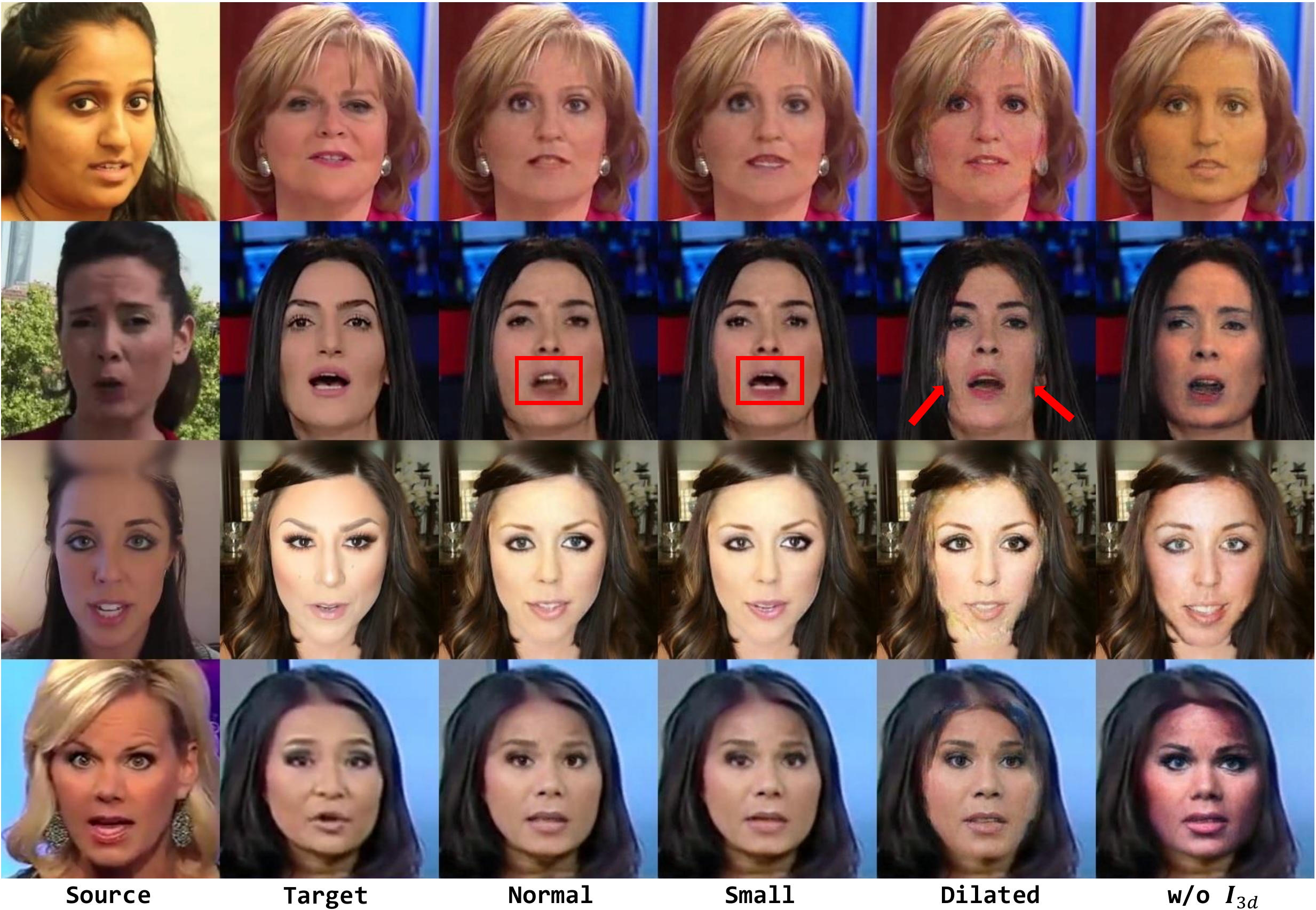}
	\caption{Qualitative ablation study of mask type and feature representation on FaceForensics++ dataset. Normal means the mask covers all face region. Small means treating the mouth area as the background based on the Normal. Dilated means dilating the Normal mask.}
	\label{fig:swap_abla}
\end{figure}

\begin{table}[t]
   \centering
   \scriptsize
   \renewcommand\arraystretch{1.2}
   \setlength\tabcolsep{4pt}
   \begin{tabular}{C{45pt}C{30pt}C{30pt}C{30pt}C{30pt}}
      \toprule
      Method & ID-A $\uparrow$ & Exp $\downarrow$  & Angle $\downarrow$ & FID $\downarrow$ \\
      \midrule
      FaceShifter & 0.5283  & \underline{2.54}   & 0.3001   & \underline{17.82}	\\
      HifiFace    & 0.5792  & 2.56   & 0.3116   & 18.91	\\
      MegaFS      & 0.3409  & 3.08   & 0.3385   & 21.68   \\ 
      InfoSwap    & \underline{0.5914}  & 2.93   & 0.2874   & 21.23	\\
      High-res    & 0.3182  & 2.92   & \underline{0.2288}   & 21.79	\\
      Ours        & \textbf{0.6121}  & \textbf{1.94}   & \textbf{0.1122}   & \textbf{15.87}	\\
      \bottomrule
   \end{tabular}
   \caption{Quantitative comparison of face swapping on FaceForensics++ dataset.}
   \label{tab:swap_sota}
\end{table}

\subsubsection{Ablation Study and Analysis}

The critical operation of our reconstruction-based face swapping paradigm is to mask the source face to avoid identity information leaking. Thus we report a visualization to explore the effect of the mask area. As depicted in Fig.~\ref{fig:swap_abla}, we design three variations, \ie, the Normal mask covers the all face area, the Small treats the mouth area as the background, and the Dilated mask dilates the Normal mask to cover more areas. There is no apparent difference between the Normal and Small types in terms of identity and attributes by comparing columns 3 and 4, but the Small obtains the more realistic mouth area since it can learn information from the Small masked source. Please pay attention to the red rectangle of row 2. The results of Dilated show the artifacts around the face contour and lead to image degradation. On the basis of these phenomenons, we choose Small mask experimentally. Besides, as shown in column 6 in Fig.~\ref{fig:swap_abla}, we observe that without the rendered face $\boldsymbol{I}_{3d}$, the color of the swapped results are prone to be similar to the source rather than the target, which further demonstrate the necessity of the rendered face as the condition.

\section{Limitations and Future Works}
First, almost all generators are based on a single image, and TGDM is no exception, which inevitably introduces
temporal inconsistency. To boost the coherence of the generated talking videos, previous works~\cite{shen2023difftalk} exploit the synthesized image as the source face for the next time step, resulting in a smoother transition between frames since the adjacent frames share the most consistent texture. However, such a frame-by-frame strategy has the problem of error propagation when encountering sudden movements, resulting in face degradation in all subsequent frames. In future work, we will be working on addressing the temporal incoherence of diffusion-based video generation.

Besides, our method retains some disadvantages of the diffusion model. For example, it takes about 45 ms on one V100 GPU to generate a single face under the $T=1000$ DDPM setting, which is unacceptable in the real application. We also do not train the model for a longer time, considering the high consumption of the diffusion model. For efficiency, our model only supports $256 \times 256$ image generation. Although DDIM~\cite{song2020denoising} and LDMs~\cite{rombach2022high} have alleviated the above problems, we hope to propose an intuitive design like StyleGAN to allow efficient high-resolution face generation.

\section{Conclusion}
In this paper, we propose a diffusion-based model to complete multimodal-driven talking face generation, which shows several appealing properties: 1) We adopt the text modal as the talking face emotion
representation, which inherits rich semantics from the CLIP, allowing flexible and generalized emotion control. 2) We treat talking face generation as a target-oriented texture transfer task. Our proposed TGDM maintains the faithful textures and undistorted appearance details from the source face and preserves explicit structural information but avoids complex texture deformations, which allows all modals to share the same generator. 3) Our proposed TGDM is also suitable for face swapping, which enables a novel reconstruction-based training paradigm and gets rid of seesaw-style optimization during inference. Our extensive results demonstrate the superiority of the proposed pipeline for various face manipulation tasks.


%





\ifCLASSOPTIONcaptionsoff
  \newpage
\fi



%

\bibliographystyle{IEEEtran}
\bibliography{main}



%








\end{document}